\documentclass{article}
\pdfoutput=1

\usepackage[preprint]{corl_2026} 

\usepackage{amsmath,amssymb,amsfonts}
\usepackage{graphicx}
\usepackage{wrapfig}
\usepackage{booktabs}
\usepackage{algorithm}
\usepackage{algorithmic}
\usepackage{xcolor}
\usepackage{multirow}
\usepackage{bm}
\usepackage{subcaption}
\usepackage{enumitem}

\newcommand{\loss}{\mathcal{L}}
\newcommand{\dataset}{\mathcal{D}}
\newcommand{\policy}{\pi_\theta}
\newcommand{\refpolicy}{\pi_{\text{ref}}}
\newcommand{\expect}{\mathbb{E}}
\newcommand{\real}{\mathbb{R}}
\newcommand{\kl}{\text{KL}}

\title{FlowPRO: Reward-Free Reinforced Fine-Tuning of Flow-Matching VLAs via Proximalized Preference Optimization}

\author{
  Yihao Wu$^{1,3}$ \quad He Zhang$^{1,2*}$ \quad Junbo Tan$^{3}$ \quad Xueqian Wang$^{3}$ \quad Zhengyou Zhang$^{1,2}$\\[0.3em]
  $^{1}$Tencent Robotics X, \quad $^{2}$Futian Laboratory, \quad $^{3}$Tsinghua University\\[0.3em]
  Project Website: \url{https://wuyeyexvnainai.github.io/flowpro/}
}

\begin{document}
\maketitle
\thispagestyle{plain} 
\renewcommand{\thefootnote}{\fnsymbol{footnote}}
\footnotetext[1]{Corresponding author}
\renewcommand{\thefootnote}{\arabic{footnote}}

\vspace{-3em}

\begin{abstract}
Post-training Vision-Language-Action (VLA) models into policies that can be reliably deployed on real robots remains a major bottleneck.
SFT and DAgger exploit failure signals only indirectly, and reward-based RL is bottlenecked by the difficulty of real-world reward design and of training reliable critics.
We present \textbf{FlowPRO}, a reward-free offline reinforced fine-tuning framework for flow-matching VLAs.
Algorithmically, we propose \textbf{RPRO} (\emph{Robotic Flow-matching Proximalized Preference Optimization}), a preference-optimization objective tailored to the flow-matching action head of VLA models. RPRO pairs a contrastive optimizer with an explicit proximal regularizer that anchors the absolute magnitude of the implicit reward, thereby eliminating the reward-hacking failure mode of plain Flow-DPO.
On the data side, a teleoperated intervention-and-rollback paradigm produces naturally paired positive and negative trajectories $(\tau^w, \tau^l)$ on a real robot from a single operator action; a Smooth Interpolation procedure, combined with batch mixing, then converts these sparse corrections into dense per-state supervision while preserving the base policy's capabilities.
On four long-horizon bimanual tasks, FlowPRO attains the highest success rate, outperforming four representative baselines, and ablations confirm the contribution of each loss component. 
\end{abstract}

\keywords{Reinforcement Learning, Preference Alignment, Vision-Language-Action Models, Flow Matching, Robot Manipulation}


\section{Introduction}
\label{sec:introduction}

Vision-Language-Action (VLA) models have recently emerged as the dominant paradigm for building generalist robots, mapping visual observations and language instructions to low-level robot actions in an end-to-end fashion~\citep{black2024pi0, brohan2023rt2, kim2024openvla, team2024octo, intelligence2025pi05, liu2024rdt, lipman2023flow, liu2023flow}.
However, post-training such models into reliably deployable policies remains challenging in practice. On real robots, both expert demonstrations and task-specific feedback signals are expensive to collect, and post-training gains on the hardest failure modes tend to saturate well before the policy reaches deployment-grade reliability~\citep{mandlekar2022matters}.

Existing real-robot post-training pipelines for VLA models fall into three families, each with a characteristic limitation that re-emerges in the flow-matching setting.
The first family, SFT and its interactive extensions---vanilla SFT~\citep{black2024pi0, kim2024openvla} and DAgger-style human correction~\citep{ross2011dagger}---scales to real hardware but only weakly exploits the failure signals from autonomous rollouts: vanilla SFT discards them entirely, while DAgger uses them merely to trigger expert correction rather than as a direct optimization signal.
The second family, reward- or value-based RL~\citep{ouyang2022training, christiano2017deep, schulman2017proximal, lu2025vlarel, guo2025iVLA, ren2024dppo, pistar06_2025}, requires training a reliable reward, value, or advantage model, which itself becomes a key obstacle for contact-rich manipulation where dense reward signals are difficult to obtain~\citep{bai2022training, levine2020offline}.
The third family, preference-based RL, bypasses reward design via preference data, with GRAPE~\citep{zhang2025grape} applying a DPO-style trajectory-level contrast over positive--negative trajectory pairs $(\tau^w, \tau^l)$~\citep{rafailov2024direct}. However, on flow-matching VLAs, the trajectory-level contrast weakens the per-state learning signal; moreover, it inherits DPO's reward-hacking failure mode~\citep{guo2025pro}, which yields implausible control actions $a$ that drift far from both the positive action $a^w$ and the negative action $a^l$.

We address these limitations with \textbf{FlowPRO}, a preference-based offline RL framework for real-robot post-training of flow-matching VLAs (Fig.~\ref{fig:overview}). FlowPRO proceeds in two stages: Stage~1 performs SFT on a task-specific dataset $\dataset_{\text{SFT}}$ to obtain a base policy from a flow-matching VLA backbone~\citep{black2024pi0,intelligence2025pi05}; Stage~2 then runs an iterative offline-RL loop on top of it for $K$ rounds.
In each round, a teleoperated intervention-and-rollback procedure (Fig.~\ref{fig:overview}(b)) lets the operator abort impending failures and teleoperate a correction from a rolled-back recovery state, producing naturally paired trajectories $(\tau^w, \tau^l)$. A Smooth Interpolation procedure (Fig.~\ref{fig:overview}(d)) then converts these sparse trajectory-level corrections into dense per-state preference tuples by synthesizing the missing counterpart at each state.
These tuples are optimized by our \textbf{RPRO} algorithm (\emph{Robotic Flow-matching Proximalized Preference Optimization}), which adopts the explicit proximal regularizer of Proximalized Preference Optimization~\citep{guo2025pro} to eliminate the reward-hacking failure mode of plain Flow-DPO, using the previous iteration's policy as the reference policy $\pi_{\text{ref}}$. We summarize our contributions as follows:

\begin{figure*}[t]
    \centering
    \includegraphics[width=0.9\textwidth]{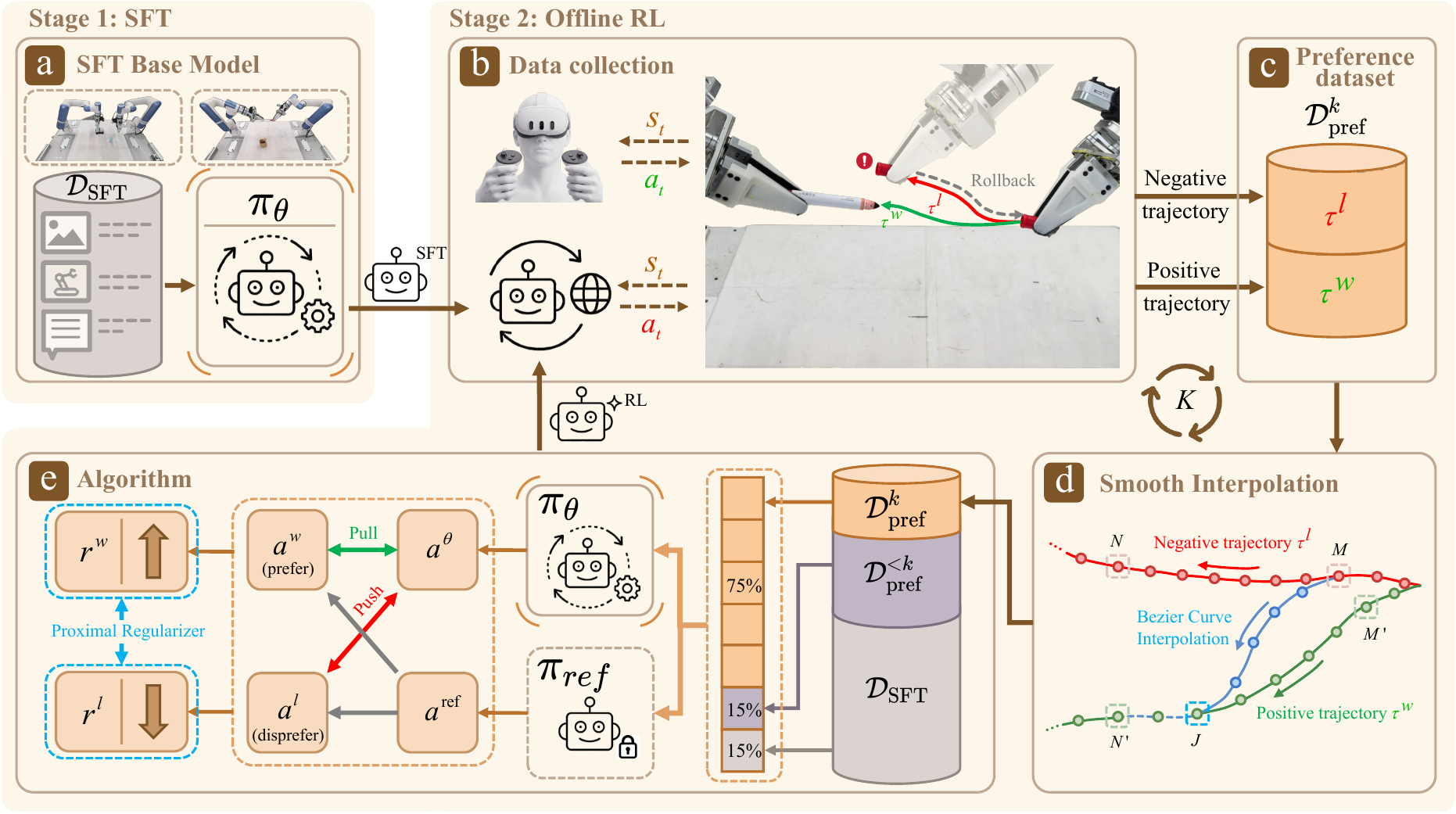}
    \caption{\textbf{Overview of the FlowPRO framework.} \textbf{(a)~SFT Base Model:} Stage~1 trains $\pi_\theta$ on $\dataset_{\text{SFT}}$. \textbf{(b)~Data Collection:} operator-triggered rollback and operator teleoperation yield paired positive and negative trajectories $\tau^w, \tau^l$. \textbf{(c)~Preference Dataset:} pairs are aggregated into $\dataset_{\text{pref}}^{k}$. \textbf{(d)~Smooth Interpolation:} B\'ezier interpolation synthesizes the missing counterpart at action-chunk granularity (e.g., $M\!\to\!J$, $J\!\to\!N'$), producing dense per-state tuples. \textbf{(e)~Algorithm (RPRO):} $a^\theta, a^{\text{ref}}$ from $\pi_\theta$ and frozen $\pi_{\text{ref}}$ are compared against $a^w/a^l$ to drive $r^w\!\uparrow, r^l\!\downarrow$, optimized on a mixed batch of $\dataset_{\text{pref}}^{k}$, $\dataset_{\text{pref}}^{<k}$, and $\dataset_{\text{SFT}}$; the updated policy is redeployed for $K$ rounds.}
    \label{fig:overview}
    \vspace{-2em}
\end{figure*}

\vspace{-0.5em}
\begin{itemize}[leftmargin=1.2em]
    \item \textbf{Algorithm (RPRO).} We introduce \textbf{RPRO}, a preference-optimization algorithm for flow-matching VLAs with two key properties: (i) an explicit proximal regularizer that eliminates the reward-hacking failure mode of plain Flow-DPO; and (ii) a gradient-vanishing property that unifies preference pairs and SFT demonstrations within a single objective.
    \item \textbf{Preference data construction.} We propose an \emph{intervention-and-rollback} paradigm for on-robot preference data collection: a single operator action yields a trajectory pair $(\tau^w, \tau^l)$, while an operator-chosen rollback horizon $\Delta$ diversifies the per-pair initial state---all without recording separate positive and negative rollouts.
    \item \textbf{Data processing and training recipe.} We design two components that make iterative RPRO fine-tuning sample-efficient under scarce real-robot corrections: a Smooth Interpolation procedure that turns sparse corrections into dense per-state preference supervision, and a fixed-ratio batch-mixing schedule that up-weights the round-$k$ dataset $\dataset_{\text{pref}}^k$ relative to the historical pool and $\dataset_{\text{SFT}}$.
\end{itemize}

\vspace{-1.5em}
\section{Related Work}
\label{sec:related_work}
\vspace{-1em}
\textbf{Reinforcement Learning for Robot Policy Fine-Tuning. }
Applying RL to imitation-pretrained robot policies has a long history. DAgger~\citep{ross2011dagger} addresses distribution shift by iteratively collecting on-policy expert corrections, and HG-DAgger~\citep{kelly2019hgdagger} introduces a human-gated variant that triggers correction only when the policy enters unsafe states; both require continuous human supervision and cannot exploit suboptimal data through a contrastive signal. On the offline-RL side, Implicit $Q$-Learning~\citep{kostrikov2022iql} and Calibrated $Q$-Learning~\citep{nakamoto2023calql} enable stable value-based fine-tuning from logged data and bridge offline pre-training with online updates. Recent VLA fine-tuning methods explore environment-reward RL~\citep{lu2025vlarel}, mixed SFT/online-RL pipelines~\citep{guo2025iVLA}, policy-gradient fine-tuning of diffusion policies (DDPO~\citep{black2024ddpo}, DPPO~\citep{ren2024dppo}), and constrained RL for safety~\citep{safevla2025}, but all of them depend on an explicit reward function or on-policy environment interaction, both of which are difficult to scale on real hardware. In contrast, our approach is fully offline and reward-free.

\textbf{Preference Optimization without Reward Models. }
\emph{Direct alignment} methods learn policies from preference data without an explicit reward model. DPO~\citep{rafailov2024direct} and its variants---KTO~\citep{ethayarajh2024kto}, IPO~\citep{azar2024general}, SimPO~\citep{meng2024simpo}, and the iterative variant of~\citet{xiong2024iterative}---all suffer from \emph{likelihood underdetermination}~\citep{guo2025pro}, which PRO~\citep{guo2025pro} resolves in the LLM setting via an explicit proximal regularizer. Contrastive Preference Learning~\citep{hejna2024cpl} additionally derives a regret-based reformulation that bypasses reward modeling for sequential decision-making. Extensions of DPO to diffusion-based generation (Diffusion-DPO~\citep{wallace2024diffusion}, D3PO~\citep{yang2024d3po}), to flow-based video generation (Flow-DPO~\citep{liu2025improving}), and to VLAs via trajectory-level contrast (GRAPE/TPO~\citep{zhang2025grape}) inherit the same likelihood-underdetermination issue, with TPO additionally diluting the per-state signal. Our work is the first to extend PRO to continuous action generation in flow-matching VLAs.

\vspace{-1em}
\section{Method}
\label{sec:method}
\vspace{-1em}
We present FlowPRO, a reward-free offline reinforced fine-tuning framework for flow-matching VLAs (overall procedure in Algorithm~\ref{alg:training}). We first review the theoretical foundations of preference optimization (\S\ref{sec:preliminaries}), then introduce RPRO, our extension of PRO to flow matching (\S\ref{sec:model_design}), and finally present our teleoperated data collection paradigm (\S\ref{sec:data_collection}) and data processing pipeline (\S\ref{sec:data_processing}).

\vspace{-1em}
\subsection{Preliminaries}
\vspace{-0.5em}
\label{sec:preliminaries}

\textbf{RLHF, DPO, and PRO. }
Standard RLHF~\citep{ouyang2022training, christiano2017deep} fits a reward model $r_\phi$ under the Bradley-Terry assumption~\citep{bradley1952rank} and optimizes a KL-regularized objective over the current policy $\policy$ and a frozen reference $\refpolicy$:
\begin{equation}
\max_\theta \; \expect_{s \sim \dataset, \, a \sim \policy(\cdot|s)} \left[ r_\phi(s, a) \right] - \beta \, \kl\!\left(\policy \| \refpolicy\right),
\label{eq:rlhf_objective}
\end{equation}
typically via PPO~\citep{schulman2017proximal}. DPO~\citep{rafailov2024direct} eliminates the explicit reward model by exploiting the closed-form mapping $r(s,a) = \beta \log \tfrac{\policy(a|s)}{\refpolicy(a|s)} + \beta \log Z(s)$ implied by Eq.~\eqref{eq:rlhf_objective}. Substituting it into the Bradley-Terry model over preferred and dispreferred action pairs $(a^w, a^l) \sim \dataset$ yields the contrastive loss
\begin{equation}
\loss_{\text{DPO}}(\theta) = -\expect_{(s, a^w, a^l) \sim \dataset} \left[ \log \sigma\!\left( \beta \log \tfrac{\policy(a^w|s)}{\refpolicy(a^w|s)} - \beta \log \tfrac{\policy(a^l|s)}{\refpolicy(a^l|s)} \right) \right].
\label{eq:dpo_loss}
\end{equation}
Since Eq.~\eqref{eq:dpo_loss} only constrains the \emph{relative} log-likelihood of $a^w$ vs.\ $a^l$, the optimizer can drive both $\policy(a^w|s)$ and $\policy(a^l|s)$ down together---a \emph{likelihood underdetermination} pathology~\citep{guo2025pro} that manifests as reward hacking even without an explicit reward model. PRO~\citep{guo2025pro} addresses this by decomposing the DPO loss into a contrastive optimizer plus a proximal regularizer anchored on a virtual \emph{hyper-response} $\mathcal{H}$ that aggregates all unobserved actions; we extend this construction to the flow-matching setting as RPRO (\S\ref{sec:model_design}).

\vspace{-1em}
\subsection{RPRO: Robotic Flow-matching Proximalized Preference Optimization}
\vspace{-0.5em}
\label{sec:model_design}
The architecture is illustrated in Fig.~\ref{fig:overview}(e).

\textbf{Preference Optimization in Flow Matching. }
In flow matching VLAs~\citep{black2024pi0}, the action head is a flow-matching model~\citep{lipman2023flow, liu2023flow}: conditioned on a state $s = (o, l)$ with visual observations $o$ and a language instruction $l$, a velocity field $v_\theta(a_t, t \mid s)$ indexed by a flow time $t \in [0, 1]$ transports Gaussian noise $\epsilon \sim \mathcal{N}(0, I)$ to an action chunk $a$ along the linear interpolant $a_t = (1-t)\epsilon + ta$, with conditional velocity $u(a_t \mid a) := a - \epsilon$. Following Flow-DPO~\citep{liu2025improving}---which applies this construction to rectified-flow video generation---and its diffusion-policy precursors~\citep{wallace2024diffusion, yang2024d3po}, we adopt the per-sample form of this regression loss as a tractable proxy for the negative log-likelihood,
\begin{equation}
\ell_\theta(s, a) = \expect_{t \sim \mathcal{U}[0,1], \, \epsilon \sim \mathcal{N}(0,I)} \big[ \| v_\theta(a_t, t \mid s) - u(a_t \mid a) \|^2 \big],
\label{eq:flow_loss}
\end{equation}
which serves as a surrogate for $-\log \policy(a \mid s)$ up to a constant, yielding the implicit-reward proxy
\begin{equation}
r_\theta(s, a) = \tfrac{\beta}{2} \big(\ell_{\text{ref}}(s, a) - \ell_\theta(s, a)\big),
\label{eq:reward_proxy}
\end{equation}
where $\ell_{\text{ref}}$ and $\ell_\theta$ denote the flow matching losses under the reference and current policies.

\textbf{PRO Loss in Flow Matching. }
Substituting Eq.~\eqref{eq:reward_proxy} into the PRO pairwise objective~\citep[Appendix~D]{guo2025pro}---DPO's contrastive log-sigmoid plus a proximal regularizer anchored on the hyper-response $\mathcal{H}$---introduces an extra hyper-response reward term $r_\theta(s, \mathcal{H})$, the implicit reward in Eq.~\eqref{eq:reward_proxy} evaluated at $\mathcal{H}$. Because the action space is continuous, the two-point set $\{a^w, a^l\}$ has Lebesgue measure zero, so $\policy(\mathcal{H} \mid s), \refpolicy(\mathcal{H} \mid s) = 1$ and $r_\theta(s, \mathcal{H}) = 0$ in the population limit (proof in Appendix~\ref{app:hyperresponse}); we therefore drop it and obtain the PRO loss in the flow-matching setting:
\begin{align}
\loss_{\text{PRO}}(\theta) = -\expect_{(s, a^w, a^l) \sim \dataset} \Big[ &\underbrace{\log \sigma\!\big(r_\theta(s, a^w) - r_\theta(s, a^l)\big)}_{\loss_{\text{con}}\text{: contrastive optimizer}} \nonumber \\
+  &\underbrace{\sum_{a \in \{a^w, a^l\}} \tfrac{1}{2} \big[ \log \sigma\!\big(r_\theta(s, a)\big) + \log \sigma\!\big(-r_\theta(s, a)\big) \big]}_{\loss_{\text{reg}}\text{: proximal regularizer}} \Big],
\label{eq:pro_loss}
\end{align}
where the regularizer is minimized at $r_\theta(s, a) = 0$ and grows symmetrically with $|r_\theta(s, a)|$, thereby anchoring the absolute magnitude of the implicit reward and preventing the likelihood underdetermination of Eq.~\eqref{eq:dpo_loss} from manifesting as reward hacking.

\textbf{Full RPRO Objective. }
To preserve base-policy performance and reinforce direct regression toward positive actions $a^w$, we combine the PRO loss with a supervised fine-tuning term weighted by loss coefficients $\lambda_{\text{PRO}}$ and $\lambda_{\text{SFT}}$:
\begin{equation}
\loss_{\text{RPRO}}(\theta) = \lambda_{\text{PRO}} \loss_{\text{PRO}}(\theta) + \lambda_{\text{SFT}} \loss_{\text{SFT}}(\theta),
\label{eq:rpro_loss}
\end{equation}
where $\loss_{\text{SFT}}(\theta) = \expect_{(s,a^w) \sim \dataset} [\ell_\theta(s, a^w)]$ is the flow-matching regression loss on positive actions.

\textbf{Contrastive Gradient Cancellation. }
Differentiating Eq.~\eqref{eq:pro_loss} yields (derivation in Appendix~\ref{app:gradient}):
\begin{align}
\nabla_\theta \loss_{\text{PRO}} = \;& \underbrace{-\sigma\!\big(r_\theta(s, a^l) - r_\theta(s, a^w)\big) \big(\nabla_\theta r_\theta(s, a^w) - \nabla_\theta r_\theta(s, a^l)\big)}_{\nabla_\theta \loss_{\text{con}}} \nonumber \\
+\;& \underbrace{\sum_{a \in \{a^w, a^l\}} \tfrac{1}{2}(2\sigma(r_\theta(s,a)) - 1) \nabla_\theta r_\theta(s,a)}_{\nabla_\theta \loss_{\text{reg}}}.
\label{eq:gradient_pro}
\end{align}
On identical-pair samples ($a^w\!=\!a^l\!=\!a$), the contrastive term cancels exactly because $\nabla_\theta r_\theta(s, a^w) - \nabla_\theta r_\theta(s, a^l) = \bm{0}$. Only the proximal regularizer survives; with $r = r_\theta(s, a)$ it reduces to $(2\sigma(r)-1)\nabla_\theta r$, which anchors $\policy$ near $\refpolicy$---analogous to the KL penalty in Eq.~\eqref{eq:rlhf_objective}. The effective RPRO gradient therefore simplifies to:
\begin{equation}
\nabla_\theta \loss_{\text{RPRO}} \big|_{a^w = a^l} = \lambda_{\text{PRO}} \nabla_\theta \loss_{\text{reg}}(\theta) + \lambda_{\text{SFT}} \nabla_\theta \loss_{\text{SFT}}(\theta),
\label{eq:gradient_vanishing}
\end{equation}
where $\loss_{\text{reg}}$ is the proximal regularizer in Eq.~\eqref{eq:pro_loss}. This cancellation makes it safe to route identical-pair samples through the same RPRO loss: the surviving regularizer prevents unconstrained drift from $\refpolicy$, while $\lambda_{\text{SFT}}$ controls the net step toward the SFT target (Appendix~\ref{app:gradient}). This property plays a key role in our data composition (\S\ref{sec:data_processing})

\vspace{-1em}
\subsection{Human-in-the-Loop Data Collection}
\vspace{-0.5em}
\label{sec:data_collection}
We collect preference trajectory pairs $(\tau^w, \tau^l)$ via a teleoperated intervention-and-rollback pipeline. Although our implementation uses Meta~Quest~3 controllers (Appendix~\ref{app:hardware}), the pipeline is device-agnostic (e.g., SpaceMouse or leader-follower arms). During rollouts of the current policy, whenever an erroneous or dangerous action is observed, the operator triggers the following sequence:
\begin{itemize}[leftmargin=1.2em]
    \item \textbf{Step 1: Rollback.} The robot rewinds to an earlier state $s_{t-\Delta}$, where the operator-chosen horizon $\Delta$ places the robot just before the erroneous behavior begins. The executed segment from $s_{t-\Delta}$ to $s_t$ is recorded as the \emph{negative trajectory} $\tau^l$.
    \item \textbf{Step 2: Scene Restoration.} The system displays the observation image at timestep $t{-}\Delta$ on the operator's VR headset, enabling the operator to reset the physical scene to $s_{t-\Delta}$.
    \item \textbf{Step 3: Human Correction.} The operator demonstrates the correct behavior from $s_{t-\Delta}$; this corrective demonstration is recorded as the \emph{positive trajectory} $\tau^w$.
\end{itemize}

This design produces naturally paired trajectories: for the same initial state $s_{t-\Delta}$, we obtain both the policy's erroneous continuation ($\tau^l$) and the human expert's correction ($\tau^w$). These pairs directly serve as preference data for RPRO training. Notably, allowing $\Delta$ to vary across interventions naturally diversifies the initial state of each pair, avoiding dependence on a fixed rollback horizon.

\vspace{-1em}
\subsection{Training Data Composition}
\label{sec:data_processing}
\vspace{-0.5em}
\textbf{Per-state pair construction.}
The RPRO loss (Eq.~\ref{eq:rpro_loss}) requires per-state tuples $(s, a^w, a^l)$. However, because $\tau^w$ and $\tau^l$ diverge after the shared initial state $s_{t-\Delta}$, each subsequent state belongs to only one trajectory---states on $\tau^l$ lack a positive action $a^w$, and states on $\tau^w$ lack a negative action $a^l$. We address this mismatch via a smooth interpolation strategy that synthesizes the missing counterpart:

\begin{itemize}[leftmargin=1.2em]
    \item \textbf{Case 1: States on the Negative Trajectory.} As illustrated in Fig.~\ref{fig:overview}(d), for a state $M$ on $\tau^l$, we construct a synthetic positive action chunk that moves from $M$ toward the corresponding segment of $\tau^w$. We first locate the closest point $M'$ on $\tau^w$ using a weighted distance metric (Appendix~\ref{app:distance_metric}) and take the next $H$ steps on $\tau^w$ as the target chunk ending at $N'$. The synthetic action $a^w$ first bridges from $M$ to a transition point $J$ on this target chunk, then follows $\tau^w$ from $J$ to $N'$. The negative action $a^l$ is the next $H$ steps along $\tau^l$ from $M$, yielding $(s_M, a^w, a^l)$.

    \item \textbf{Case 2: States on the Positive Trajectory.} For a state on $\tau^w$, the positive action $a^w$ is the next $H$ steps along $\tau^w$, and we set $a^l = a^w$. By contrastive gradient cancellation (Eq.~\ref{eq:gradient_vanishing}), only the SFT loss and proximal regularizer remain active, both reinforcing correct behavior.

    \item \textbf{Case 3: SFT Demonstration Data.} For states from $\dataset_{\text{SFT}}$, we similarly set $a^w = a^l = a_{\text{SFT}}$. As in Case~2, contrastive gradients vanish; the proximal regularizer anchors $\policy$ toward $\refpolicy$, rendering these regularized SFT samples.
\end{itemize}

Case~1 implements this bridge with a cubic B\'ezier curve for position, Slerp for orientation, and linear interpolation for the gripper. The B\'ezier curve uses only the arrival tangent from $\tau^w$ at $J$, so the synthetic action merges smoothly into $\tau^w$ without inheriting the erroneous direction of $\tau^l$ at $M$ (Appendix~\ref{app:interpolation}). We analyze physical-plausibility safeguards in Appendix~\ref{app:physical_plausibility}.

The combined dataset takes the following form:
\begin{align}
\dataset = \Big\{ &(s, a^w, a^l) \,\Big|\, s \in \tau^l: a^w = \text{Interp}(M \to N'), \; a^l = \tau^l(M \to N) \Big\} \nonumber \\
\cup \; \Big\{ &(s, a^w, a^l) \,\Big|\, s \in \tau^w: a^w = a^l = \tau^w(s \to s+H) \Big\} \nonumber \\
\cup \; \Big\{ &(s, a^w, a^l) \,\Big|\, s \in \dataset_{\text{SFT}}: a^w = a^l = a_{\text{SFT}} \Big\}.
\label{eq:dataset_format}
\vspace{-0.5em}
\end{align}

\textbf{Batch mixing across iterations.}
Let $\dataset_{\text{pref}}^k$ be the preference dataset built via Smooth Interpolation in RPRO iteration $k$, and $\dataset_{\text{pref}}^{<k} \triangleq \bigcup_{j<k}\dataset_{\text{pref}}^j$ the historical preference pool. We keep these three sources in separate buffers and form each mini-batch $\mathcal{B}$ by drawing samples from them at fixed proportions:
\[
k{=}1:\ 80\%\,\text{from}\,\dataset_{\text{pref}}^k,\ 20\%\,\text{from}\,\dataset_{\text{SFT}}; \quad k{\geq}2:\ 70\%\,\text{from}\,\dataset_{\text{pref}}^k,\ 15\%\,\text{from}\,\dataset_{\text{pref}}^{<k},\ 15\%\,\text{from}\,\dataset_{\text{SFT}}.
\]

This fixed-ratio schedule balances three goals. The current preference set $\dataset_{\text{pref}}^k$ receives the largest share because it contains the newest failure states and thus the most targeted corrective signal. The historical pool $\dataset_{\text{pref}}^{<k}$ is replayed from $k{\geq}2$ onward to prevent regression on previously corrected failures. The SFT share preserves base-policy capabilities and mitigates catastrophic forgetting.


\vspace{-1em}
\section{Experimental Evaluation}
\label{sec:experiments}
\vspace{-1em}

\begin{figure*}[t]
    \centering
    \includegraphics[width=\textwidth]{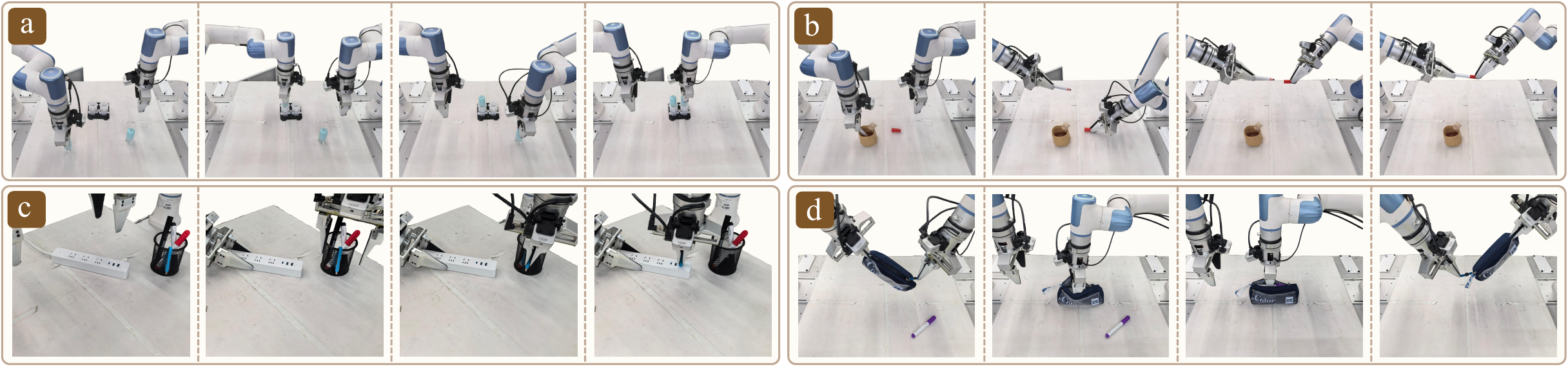}
    \caption{\textbf{Experimental setup for the four real-robot tasks:} \textbf{(a)} cosmetic packaging (\textsc{Pack}), \textbf{(b)} pen-cap assembly (\textsc{Cap}), \textbf{(c)} USB insertion (\textsc{USB}), and \textbf{(d)} pencil-case packing (\textsc{Case}). Due to space constraints, this figure shows only a simplified overview; the complete per-stage pipeline for each task is provided in Fig.~\ref{fig:complete_experiment_task} of Appendix~\ref{app:tasks}.}
    \label{fig:experiment_task}
    \vspace{-1.5em}
\end{figure*}

We evaluate FlowPRO along four axes:
\textbf{(Q1)} \emph{Overall performance:} Does FlowPRO surpass representative baselines from both the \emph{positive-only} and the \emph{positive-and-negative} regimes?
\textbf{(Q2)} \emph{Data construction:} Does state-wise \emph{Smooth Interpolation} outperform trajectory-wise preference contrast?
\textbf{(Q3)} \emph{Loss design:} Does the proximal regularizer in RPRO eliminate the reward-hacking failure mode of plain Flow-DPO?
\textbf{(Q4)} \emph{Component attribution:} How do the SFT term and the proximal regularizer individually contribute to the full RPRO objective?

\vspace{-1em}
\subsection{Experimental Setup}
\label{sec:tasks}
\vspace{-0.5em}
\textbf{Hardware, Tasks, and Protocol. }
\textit{Platform.} All experiments are conducted on a Dobot XTrainer bimanual platform (Appendix~\ref{app:hardware}). \textit{Tasks.} We evaluate on four long-horizon bimanual tasks (Fig.~\ref{fig:experiment_task}, Appendix~\ref{app:tasks}): \textsc{Pack} (sub-cm insertion), \textsc{Cap} (in-air coordination), \textsc{USB} (sub-mm precision), and \textsc{Case} (deformable long-horizon packing). \textit{Training.} We first train a base SFT policy on $\dataset_{\text{SFT}}$, which serves as the reference policy $\pi_{\text{ref}}$, and then perform $K=3$ rounds of RPRO fine-tuning on $\{\dataset_{\text{pref}}^k\}_{k=1}^{K}$. \textit{Evaluation.} To control for training-time stochasticity, each entry in Table~\ref{tab:main_results} is obtained from 3 independent training seeds, and we report the cross-seed mean~$\pm$~standard deviation, where each per-seed success rate is computed over $n{=}100$ rollouts with randomized initial placements.

\textbf{Baselines and Ablations. }
\label{sec:baselines}%
\textit{Baselines (Q1, Q2).} We compare FlowPRO against four representative comparators covering both the \emph{positive-only} and the \emph{positive-and-negative} regimes: \textbf{DAgger}~\citep{ross2011dagger} (dataset aggregation), \textbf{DAgger-Buffered} (a positive-only control that isolates the preference loss from our batch-composition schedule), \textbf{PI0.6*}~\citep{pistar06_2025} (advantage conditioning), and \textbf{TPO}~\citep{zhang2025grape} (trajectory-wise preference optimization). All comparators start from the same SFT checkpoint and follow the same iterative data-collection protocol. \textit{Ablations (Q3, Q4).} We further ablate our loss along the path from a preference-free objective to the full design: \textbf{SFT}, \textbf{DPO} (Eq.~\ref{eq:dpo_loss} applied to flow matching, without the SFT term or the proximal regularizer), \textbf{DPO+SFT}, \textbf{PRO} (RPRO without the SFT term, i.e., $\loss_{\text{PRO}}$ alone), and \textbf{RPRO} (the full objective in Eq.~\ref{eq:rpro_loss}). To manage real-robot cost, ablations are run \emph{only on \textsc{Pack}} (which exhibits all representative failure modes) and evaluated after a single round of fine-tuning. We defer the detailed baseline descriptions and the exact loss expressions of every variant to Appendices~\ref{app:baseline_details} and~\ref{app:ablation_losses}.

\vspace{-1em}
\subsection{Experimental Results}
\label{sec:results}
\vspace{-0.5em}

\begin{table}[t]
\centering
\caption{\textbf{Final success rate and completion time after $K{=}3$ rounds of iterative post-training on four real-robot bimanual tasks.} SR ($\uparrow$, \%) is reported as mean~$\pm$~std (in pp) across 3 training seeds, with each per-seed SR computed over $n{=}100$ randomized rollouts; CT ($\downarrow$, s) is reported as the mean over the same rollouts. 
}
\label{tab:main_results}
\setlength{\tabcolsep}{4pt}
\begin{tabular*}{\textwidth}{@{\extracolsep{\fill}}clcccccccc@{}}
\toprule
\multirow{2}{*}{\begin{tabular}[c]{@{}c@{}}Base\\ Policy\end{tabular}} & \multicolumn{1}{c}{\multirow{2}{*}{Fine-tune}} & \multicolumn{2}{c}{\textsc{Pack}} & \multicolumn{2}{c}{\textsc{Cap}} & \multicolumn{2}{c}{\textsc{USB}} & \multicolumn{2}{c}{\textsc{Case}} \\ \cmidrule(lr){3-4} \cmidrule(lr){5-6} \cmidrule(lr){7-8} \cmidrule(lr){9-10}
                                                                       & \multicolumn{1}{c}{}                           & SR   & CT  & SR   & CT  & SR   & CT  & SR   & CT  \\ \midrule
\multirow{5}{*}{\textsc{PI0}}                                          & Dagger        & $88\!\pm\!1.3$\% & 32s & $83\!\pm\!3.0$\% & 31s & $80\!\pm\!2.8$\% & 35s & $83\!\pm\!2.6$\% & 58s \\
                                                                       & Dagger-Buffer & $90\!\pm\!2.0$\% & 27s & $89\!\pm\!2.4$\% & 29s & $85\!\pm\!1.6$\% & 33s & $89\!\pm\!2.2$\% & 57s \\
                                                                       & PI0.6*        & $94\!\pm\!1.5$\% & 25s & $91\!\pm\!1.8$\% & 28s & $90\!\pm\!3.2$\% & 27s & $90\!\pm\!1.4$\% & 46s \\
                                                                       & TPO           & $92\!\pm\!1.7$\% & 28s & $94\!\pm\!2.2$\% & 27s & $87\!\pm\!1.7$\% & 30s & $88\!\pm\!2.3$\% & 54s \\
                                                                       & \textbf{RPRO} & $\bm{99\!\pm\!1.0}$\textbf{\%} & \textbf{20s} & $\bm{98\!\pm\!1.5}$\textbf{\%} & \textbf{24s} & $\bm{92\!\pm\!1.5}$\textbf{\%} & \textbf{26s} & $\bm{93\!\pm\!2.0}$\textbf{\%} & \textbf{42s} \\ \midrule
\multirow{5}{*}{\textsc{PI0.5}}                                        & Dagger        & $86\!\pm\!2.5$\% & 30s & $86\!\pm\!2.3$\% & 31s & $82\!\pm\!3.4$\% & 28s & $85\!\pm\!3.3$\% & 57s \\
                                                                       & Dagger-Buffer & $93\!\pm\!1.3$\% & 28s & $92\!\pm\!1.5$\% & 28s & $89\!\pm\!2.1$\% & 25s & $87\!\pm\!2.4$\% & 55s \\
                                                                       & PI0.6*        & $92\!\pm\!2.8$\% & 23s & $94\!\pm\!1.3$\% & 27s & $93\!\pm\!1.9$\% & 26s & $88\!\pm\!2.7$\% & 43s \\
                                                                       & TPO           & $95\!\pm\!1.6$\% & 24s & $93\!\pm\!2.8$\% & 26s & $91\!\pm\!2.4$\% & 27s & $90\!\pm\!1.5$\% & 45s \\
                                                                       & \textbf{RPRO} & $\bm{99\!\pm\!1.0}$\textbf{\%} & \textbf{17s} & $\bm{99\!\pm\!1.0}$\textbf{\%} & \textbf{23s} & $\bm{95\!\pm\!1.6}$\textbf{\%} & \textbf{24s} & $\bm{93\!\pm\!1.2}$\textbf{\%} & \textbf{38s} \\ \bottomrule
\end{tabular*}
\vspace{-1em}
\end{table}

\begin{figure*}[t]
    \centering
    \includegraphics[width=\textwidth]{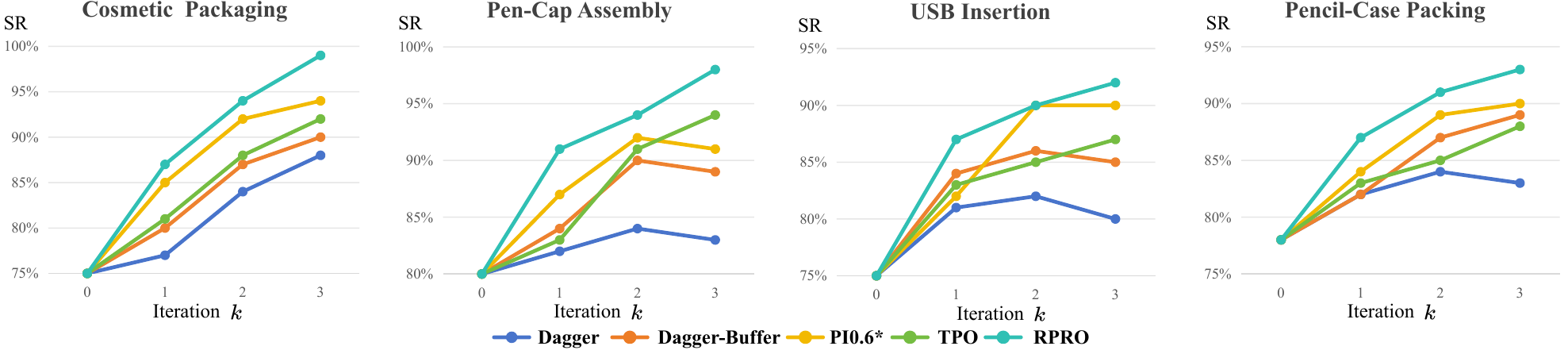}
    \caption{\textbf{Per-iteration success rates (SR) on the four real-robot tasks, with \textsc{PI0.5} as the base policy.} Iteration~$0$ corresponds to the shared SFT checkpoint. The corresponding curves with \textsc{PI0} as the base policy are reported in Fig.~\ref{fig:baseline_curves_pi0} of Appendix~\ref{app:baseline_details}.}
    \label{fig:baseline_curves}
    \vspace{-2em}
\end{figure*}
\begin{figure*}[t]
    \centering
    \includegraphics[width=\textwidth]{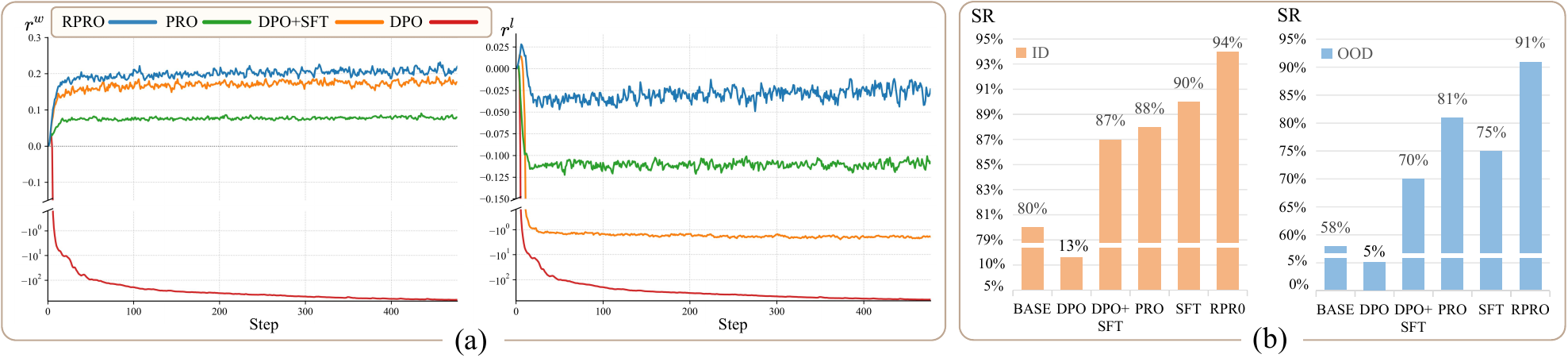}
    \caption{\textbf{Loss-component ablations of RPRO.} \textbf{(a)} Implicit-reward dynamics during training: per-step rewards on positive actions $r^{w}$ (\emph{left}) and negative actions $r^{l}$ (\emph{right}) for RPRO, PRO, DPO+SFT, and DPO. \textbf{(b)} Task success rates under in-distribution (\emph{left}) and out-of-distribution (\emph{right}) initial conditions across loss-component variants. Here, \emph{out-of-distribution (OOD)} refers to object initial positions sampled from spatial regions that were never visited during SFT training.}
    \label{fig:training_curves}
    \vspace{-2em}
\end{figure*}

Final and per-iteration success rates on \textsc{Pack}, \textsc{Cap}, \textsc{USB}, and \textsc{Case} are reported in Table~\ref{tab:main_results} and Fig.~\ref{fig:baseline_curves}, and loss-component ablations of RPRO are shown in Fig.~\ref{fig:training_curves}. 
RPRO strictly dominates each baseline across all 8 task--base strata, with a Bonferroni-corrected one-sided Cochran--Mantel--Haenszel (CMH) test yielding $p<10^{-3}$ against each of DAgger, DAgger-Buffered, PI0.6*, and TPO (Appendix~\ref{app:significance}).
RPRO likewise strictly dominates each loss-component variant on \textsc{Pack} under in-distribution (ID) and out-of-distribution (OOD) initial conditions, with the same CMH test yielding $p \leq 6\times 10^{-3}$ against each of PRO, SFT, DPO+SFT, and DPO (Appendix~\ref{app:sig_ablation}). We discuss the findings below by question.



\textbf{Comparison with baselines (Q1 \& Q2).}

\textit{RPRO vs.~TPO.}
Our state-wise method outperforms the trajectory-wise TPO variant by 3--7\,pp on every stratum (Table~\ref{tab:main_results}), directly addressing Q2 (state-wise vs.\ trajectory-wise contrast). Trajectory-level contrast only constrains the sum of per-step log-ratios, so the per-state learning signal is diluted by the trajectory length; in contrast, our per-state Smooth Interpolation gives every training state a well-defined pairwise signal at action-chunk granularity. Trajectory-level contrast is also memory-intensive: keeping a full trajectory in memory forces sub-sampling of states on $\tau^w$ and $\tau^l$, so the positive trajectory is only partially learned.

\textit{RPRO vs.~PI0.6*.}
RPRO outperforms the advantage-conditioned PI0.6* baseline by 2--7\,pp on \emph{identical} preference data, isolating \emph{conditioning}-based from \emph{contrastive-loss}-based use of the signal. PI0.6* relies on the model to discover the ``improved''/``unimproved'' partition from a single conditioning token under a pure regression objective---an indirect pressure that can be diluted by the rest of the VLM context---whereas RPRO injects the preference signal directly into the action-generation loss, pushing $\pi_\theta$ toward $a^w$ and away from $a^l$ per state and per chunk.

\textit{RPRO vs.~DAgger \& DAgger-Buffered.}
Both DAgger variants rely on positive samples only, while RPRO additionally exploits negative trajectories through contrastive learning, yielding a stronger learning signal and an 8--15\,pp gain over vanilla DAgger across strata. Notably, DAgger-Buffered alone outperforms vanilla DAgger by 2--7\,pp, confirming that our batch-composition schedule is a more sample-efficient way to consume the round-$k$ corrections than merging them uniformly into $\dataset_{\text{SFT}}$: corrections form a small but informative set of states where the current policy fails, and therefore deserve heavier per-batch weighting.

\textbf{Loss-component ablations (Q3 \& Q4).}

\textit{RPRO vs.~PRO} (ablates $\loss_{\text{SFT}}$).
Adding the SFT term to PRO yields a clear performance improvement, for two reasons. \emph{First}, the proximal regularizer prevents collapse but does not actively pull $\pi_\theta$ toward $a^w$ beyond maintaining a positive log-ratio gap $r^w > r^l$; the SFT term fills this gap by explicitly fitting $\pi_\theta$ to $a^w$ in the flow-matching regression sense, accelerating convergence. \emph{Second}, the SFT term continually reinforces the base SFT skills, mitigating catastrophic forgetting.
SR drops from $94\%/91\%$ to $88\%/81\%$ (ID/OOD).

\textit{RPRO vs.~SFT} (ablates $\loss_{\text{PRO}}$).
Adding the contrastive term in $\loss_{\text{PRO}}$ to SFT clearly improves performance, especially under OOD initial conditions. SFT alone only learns to \emph{imitate} positive actions and never observes what \emph{not} to do, whereas RPRO exposes the policy to both positive and negative actions, providing an explicit \emph{push-away} gradient from $a^l$ that pulls the policy back from nearby failure modes; the OOD-stratum gap is the dominant driver of RPRO's overall gain over SFT.
SR drops to $90\%/75\%$.

\textit{RPRO vs.~DPO+SFT} (ablates $\loss_{\text{reg}}$).
This comparison isolates the proximal regularizer $\loss_{\text{reg}}$, with the SFT anchor fixed in both variants. 
Since the SFT term is \emph{asymmetric}---anchoring $\pi_\theta$ only at $a^w$---DPO's contrastive optimizer drives $r^l$ unchecked into an unbounded regime, while the SFT pull keeps $r^w$ deceptively high (Fig.~\ref{fig:training_curves}(a)). Such \emph{unilateral runaway} on the rejected side, even when the margin $r^w - r^l$ remains positive, contaminates the action distribution $\pi_\theta$ rolls out at inference and degrades execution.
SR drops to $87\%/70\%$.

\textit{RPRO vs.~DPO} (ablates $\loss_{\text{reg}} + \loss_{\text{SFT}}$).
Removing both the SFT anchor and the proximal regularizer turns the loss into plain DPO, which exhibits the reward-hacking failure mode in its most extreme form. In Fig.~\ref{fig:training_curves}(a), $r^w$ itself drives below zero and continues to fall to the $-10^2$ scale---i.e., $\pi_\theta$ ends up \emph{farther} from $a^w$ than the reference policy in absolute terms, even though the relative gap $r^w - r^l$ stays positive throughout training. RPRO's proximal regularizer eliminates this pathology by anchoring the absolute reward levels of both $a^w$ and $a^l$ around $\pi_{\text{ref}}$.
SR collapses to $13\%/5\%$.


\vspace{-1em}
\section{Conclusion}
\label{sec:conclusion}
\vspace{-1em}
In this paper, we presented \textbf{FlowPRO}, a reward-free offline reinforced fine-tuning framework for flow-matching VLAs, built around the \textbf{RPRO} loss and a teleoperated intervention-and-rollback paradigm for preference data construction. 
Across four long-horizon bimanual real-robot tasks and two $\pi_0$-family base policies, FlowPRO consistently attains the highest success rate and the shortest completion time, outperforming all four representative baselines and every loss-component ablation; diagnostic implicit-reward curves further show that the proximal regularizer cleanly removes the reward-hacking failure mode exhibited by Flow-DPO.
These results indicate that FlowPRO is an effective and reliable post-training recipe for flow-matching VLA policies. With only a small number of human interventions, it converts scarce on-robot corrections into deployment-grade policy improvements.

\vspace{-1em}
\section{Limitations and Future Work}
\label{sec:limitations}
\vspace{-1em}
Our study has several limitations, each pointing to a concrete extension.
First, our evaluation is restricted to a single bimanual platform; broader generalization to mobile and dexterous manipulation is left as future work.
Second, deciding \emph{when} to roll back and \emph{how far} (the rollback horizon $\Delta$) currently relies on a human operator. A natural extension is a learned failure-detector that triggers rollback autonomously, leaving the human only the correction.


\bibliography{references}


\clearpage
\appendix
\section{Asymptotic Vanishing of the Hyper-Response Reward in Continuous Action Spaces}
\label{app:hyperresponse}

This appendix provides the formal asymptotic argument supporting the simplification $r_\theta(s, \mathcal{H}) \approx 0$ used in Eq.~\eqref{eq:pro_loss}. In the original PRO derivation~\citep{guo2025pro}, the hyper-response $\mathcal{H}$ aggregates all responses not in the observed preference pair, and is treated under the discrete vocabulary of an LLM. Here we restate the argument under the continuous-action regime of flow-matching VLAs and show that, in this regime, the hyper-response reward is not merely small but \emph{vanishes exactly} in the population limit.

\paragraph{Setup and notation.}
Fix a state $s = (o, l)$ and let $\mathcal{A} \subseteq \real^d$ denote the (continuous) action-chunk space. We treat $\pi_\theta(\cdot \mid s)$ and $\pi_{\text{ref}}(\cdot \mid s)$ as absolutely-continuous probability measures on $\mathcal{A}$ with densities $p_\theta(a \mid s)$ and $p_{\text{ref}}(a \mid s)$ w.r.t.\ the Lebesgue measure $\mu$. For an observed preference pair $(a^w, a^l) \in \mathcal{A}^2$ at state $s$, define the \emph{hyper-response set}
\begin{equation}
\mathcal{H} \;\triangleq\; \mathcal{A} \setminus \{a^w, a^l\},
\label{eq:H_def_continuous}
\end{equation}
i.e., the (uncountable) set of all action chunks at state $s$ other than the two observed responses. Following PRO~\citep[Appendix~D]{guo2025pro}, the hyper-response reward is
\begin{equation}
r_\theta(s, \mathcal{H}) \;\triangleq\; \beta \, \log \frac{\pi_\theta(\mathcal{H} \mid s)}{\pi_{\text{ref}}(\mathcal{H} \mid s)},
\label{eq:rH_def}
\end{equation}
which is the analogue of $r_\theta(s, a) = \beta \log \tfrac{\pi_\theta(a \mid s)}{\pi_{\text{ref}}(a \mid s)}$ but evaluated on a set rather than a singleton (note that we use $\beta$ rather than $\tfrac{\beta}{2}$ here to match the LLM-PRO convention; the flow-matching surrogate Eq.~\ref{eq:reward_proxy} carries the same $\tfrac{\beta}{2}$ scaling on both numerator and denominator, so this choice is immaterial for the present argument).

\paragraph{Proposition (exact vanishing).}
Let $\pi_\theta(\cdot \mid s)$ and $\pi_{\text{ref}}(\cdot \mid s)$ be absolutely continuous w.r.t.\ Lebesgue measure on $\mathcal{A}$. Then for any finite preference pair $(a^w, a^l) \in \mathcal{A}^2$,
\begin{equation}
\pi_\theta(\mathcal{H} \mid s) \;=\; \pi_{\text{ref}}(\mathcal{H} \mid s) \;=\; 1, \qquad\text{hence}\qquad r_\theta(s, \mathcal{H}) \;=\; 0.
\label{eq:rH_zero}
\end{equation}

\paragraph{Proof.}
The complement of $\mathcal{H}$ in $\mathcal{A}$ is the two-point set $\{a^w, a^l\}$, which has Lebesgue measure zero:
\begin{equation}
\mu(\{a^w, a^l\}) \;=\; \mu(\{a^w\}) + \mu(\{a^l\}) \;=\; 0 + 0 \;=\; 0.
\end{equation}
Since $\pi_\theta(\cdot \mid s)$ admits a density w.r.t.\ $\mu$, absolute continuity gives
\begin{equation}
\pi_\theta(\{a^w, a^l\} \mid s) \;=\; \int_{\{a^w, a^l\}} p_\theta(a \mid s) \, d\mu(a) \;=\; 0,
\end{equation}
and therefore $\pi_\theta(\mathcal{H} \mid s) = 1 - \pi_\theta(\{a^w, a^l\} \mid s) = 1$. The same argument applies to $\pi_{\text{ref}}(\mathcal{H} \mid s)$. Plugging both into Eq.~\eqref{eq:rH_def} gives $r_\theta(s, \mathcal{H}) = \beta \log 1 = 0$. \hfill$\square$

\paragraph{Why the discrete-action argument is not directly applicable.}
In the LLM setting of~\citet{guo2025pro}, $\mathcal{A}$ is a finite vocabulary $\mathcal{V}^L$ of token sequences and $\pi_\theta(\cdot\mid s)$ is a probability \emph{mass} function. There, $\pi_\theta(\{a^w, a^l\}\mid s) = p_\theta(a^w\mid s) + p_\theta(a^l\mid s)$ is generally \emph{nonzero}, so $\pi_\theta(\mathcal{H}\mid s) = 1 - p_\theta(a^w\mid s) - p_\theta(a^l\mid s)$. The approximation $r_\theta(s,\mathcal{H}) \approx 0$ in PRO~\citep{guo2025pro} relies on the heuristic that $p_\theta(a^w\mid s), p_\theta(a^l\mid s) \ll 1$ when the vocabulary is large, so $\pi_\theta(\mathcal{H}\mid s)\to 1$ as $|\mathcal{V}^L|\to\infty$. The continuous-action regime is qualitatively stronger: \emph{individual} actions carry zero probability mass under any absolutely-continuous density, so the simplification $r_\theta(s, \mathcal{H}) = 0$ is \emph{exact} rather than asymptotic. The flow-matching architecture of $\pi_0$~\citep{black2024pi0}, which generates continuous action chunks via a velocity-field ODE, satisfies this absolute-continuity assumption by construction whenever the noise prior $\mathcal{N}(0, I)$ is full-rank---which it is.

\paragraph{Robustness to finite-precision and stochastic-sampler effects.}
A pedantic objection is that in practice $\pi_\theta$ is realised by a stochastic flow-matching sampler in finite floating-point precision, so the realised distribution is technically supported on a finite grid. In this case the proposition above does not apply verbatim. We address this in two ways. (i)~The grid spacing $\delta$ induced by floating-point precision is $\sim 10^{-7}$ in fp32; for an action chunk of dimension $d \cdot H$ (e.g., $d = 10$ features per arm---xyz, 6D~rotation, and gripper width---times $2$ arms, $H = 50$ steps, so $\text{dim} = 20 \cdot 50 = 1000$ for bimanual chunks), the probability mass concentrated on the singleton $\{a^w\}$ relative to the support of $\pi_\theta$ scales as $\delta^{\text{dim}}$, which is vanishingly small. (ii)~More importantly, in the training loss we never need to evaluate $\pi_\theta(\{a^w\}\mid s)$ directly; we only need $\ell_\theta(s, a^w)$, the flow-matching loss surrogate (see \S\ref{sec:model_design}). The surrogate is a continuous functional that is well-defined regardless of whether we view $\pi_\theta$ as truly continuous or as a fine-grid approximation. Consequently the argument carries over with quantitative slack $|r_\theta(s,\mathcal{H})| \le O(\delta^{\text{dim}})$, which is many orders of magnitude below the typical implicit-reward scales we observe in Fig.~\ref{fig:training_curves}(a).


\paragraph{Summary.}
Under the absolute-continuity assumption that holds by construction for flow-matching VLA heads with a full-rank Gaussian prior, the hyper-response reward $r_\theta(s, \mathcal{H})$ is \emph{identically zero}, not merely $\ll 1$. This sharpens the LLM-PRO heuristic and justifies dropping the $r_\theta(s, \mathcal{H})$ term from the contrastive-plus-regularizer objective in Eq.~\eqref{eq:pro_loss}. The remaining loss is exact---not approximate---under the stated assumption.

\section{Derivation of the Gradient Property for Identical Pairs}
\label{app:gradient}

We derive the gradient of $\loss_{\text{PRO}}$ (Eq.~\ref{eq:pro_loss}) and analyze its behavior when $a^w = a^l$.

\paragraph{Step 1: General gradient.}
The PRO loss is:
\begin{align}
\loss_{\text{PRO}} = -\expect_{(s, a^w, a^l)} \Big[ &\log \sigma\!\big(r_\theta(s, a^w) - r_\theta(s, a^l)\big) \nonumber \\
+ \sum_{a \in \{a^w, a^l\}} \Big( &\tfrac{1}{2} \log \sigma\!\big(r_\theta(s, a)\big) + \tfrac{1}{2} \log \sigma\!\big(-r_\theta(s, a)\big) \Big) \Big].
\end{align}
Using $\nabla_x \log\sigma(x) = \sigma(-x) = 1 - \sigma(x)$ and the outer negative sign, we differentiate each term with respect to $\theta$:

\textbf{Contrastive term.} Let $\Delta = r_\theta(s, a^w) - r_\theta(s, a^l)$:
\begin{equation}
-\nabla_\theta \log\sigma(\Delta) = -\sigma(-\Delta) \cdot (\nabla_\theta r_\theta(s, a^w) - \nabla_\theta r_\theta(s, a^l)).
\end{equation}

\textbf{Regularizer term.} For each $a \in \{a^w, a^l\}$, let $r = r_\theta(s, a)$:
\begin{align}
-\nabla_\theta \Big[\tfrac{1}{2}\log\sigma(r) + \tfrac{1}{2}\log\sigma(-r)\Big] &= -\tfrac{1}{2}(1 - 2\sigma(r)) \cdot \nabla_\theta r \nonumber \\
&= \tfrac{1}{2}(2\sigma(r) - 1) \cdot \nabla_\theta r.
\end{align}

Combining the contrastive and regularizer contributions yields the full gradient:
\begin{align}
\nabla_\theta \loss_{\text{PRO}} = \;& \underbrace{-\sigma\!\big(r_\theta(s, a^l) - r_\theta(s, a^w)\big) \cdot \big(\nabla_\theta r_\theta(s, a^w) - \nabla_\theta r_\theta(s, a^l)\big)}_{\text{contrastive optimizer}} \nonumber \\
+\;& \underbrace{\sum_{a \in \{a^w, a^l\}} \tfrac{1}{2}(2\sigma(r_\theta(s,a)) - 1) \cdot \nabla_\theta r_\theta(s,a)}_{\text{proximal regularizer}}.
\label{eq:gradient_pro_app}
\end{align}

\paragraph{Step 2: Specialization to $a^w = a^l = a$.}
When $a^w = a^l = a$, we have $r_\theta(s, a^w) = r_\theta(s, a^l) \triangleq r$, so $\Delta = 0$.

\textbf{Contrastive term}: $\nabla_\theta r_\theta(s, a^w) - \nabla_\theta r_\theta(s, a^l) = \bm{0}$, so this term vanishes \emph{automatically}, regardless of the prefactor $-\sigma(0) = -\frac{1}{2}$---no indicator function or sample filtering is required.

\textbf{Regularizer terms}: The two terms (for $a^w$ and $a^l$) are identical, each equal to $\frac{1}{2}(2\sigma(r)-1) \cdot \nabla_\theta r$. They sum to $(2\sigma(r)-1) \cdot \nabla_\theta r$, which acts as a safeguard preventing $|r|$ from growing unbounded (equivalently, preventing $\pi_\theta$ from drifting too far away from $\pi_{\text{ref}}$)---playing the role of a trust-region / KL-type penalty analogous to the regularizers used in RLHF.

\paragraph{Step 3: The surviving regularizer as a trust-region anchor.}
The residual regularizer gradient $(2\sigma(r)-1)\cdot\nabla_\theta r$ plays the exact role of a trust-region / KL-type penalty. Recall $r = r_\theta(s, a) = \tfrac{\beta}{2}(\ell_{\text{ref}}(s, a) - \ell_\theta(s, a))$, so by the chain rule
\begin{equation}
\nabla_\theta r_\theta(s, a) = -\tfrac{\beta}{2}\,\nabla_\theta \ell_\theta(s, a),
\label{eq:grad_r_vs_sft}
\end{equation}
which is, up to the positive scalar $\tfrac{\beta}{2}$, the \emph{negative} SFT gradient on $(s, a)$. Substituting into the regularizer gradient gives
\begin{equation}
(2\sigma(r)-1)\cdot\nabla_\theta r_\theta(s, a) \;=\; -\tfrac{\beta}{2}(2\sigma(r)-1)\cdot\nabla_\theta \ell_\theta(s, a).
\label{eq:reg_grad_chain}
\end{equation}
Two useful consequences follow.

\emph{(a)~Controlled opposition to SFT---the hallmark of a trust-region penalty.} The SFT term on $(s, a)$ performs gradient descent on $\ell_\theta(s, a)$, i.e.\ it updates $\theta$ along $-\nabla_\theta \ell_\theta(s, a)$ to make $\pi_\theta$ better reproduce $a$. The regularizer's descent direction, by Eq.~\eqref{eq:reg_grad_chain}, is
\begin{equation}
-\nabla_\theta \loss_{\text{reg}} \;=\; -\tfrac{1}{2}(2\sigma(r)-1)\,\nabla_\theta r_\theta(s, a) \;=\; +\tfrac{\beta}{4}(2\sigma(r)-1)\,\nabla_\theta \ell_\theta(s, a).
\label{eq:reg_descent}
\end{equation}
From Eq.~\eqref{eq:reg_descent}, the sign of the regularizer's descent step relative to the SFT descent step $-\nabla_\theta \ell_\theta(s, a)$ is determined entirely by the sign of $r$, and the two regimes have to be analyzed separately.

\begin{itemize}[leftmargin=1.2em]
    \item \emph{Case $r > 0$} (i.e.\ $\pi_\theta$ has reached a \emph{higher} likelihood for $a$ than $\pi_{\text{ref}}$, $\ell_\theta(s, a) < \ell_{\text{ref}}(s, a)$, so $2\sigma(r)-1 > 0$). The regularizer's descent step is a positive multiple of $+\nabla_\theta \ell_\theta(s, a)$, i.e.\ \emph{anti-parallel} to the SFT descent step. It acts as a mild brake, pulling $\ell_\theta(s, a)$ back up and preventing $\pi_\theta(a \mid s)$ from running too far above $\pi_{\text{ref}}(a \mid s)$.

    \item \emph{Case $r < 0$} (i.e.\ $\pi_\theta$ has a \emph{lower} likelihood for $a$ than $\pi_{\text{ref}}$, $\ell_\theta(s, a) > \ell_{\text{ref}}(s, a)$, so $2\sigma(r)-1 < 0$). The regularizer's descent step is now a positive multiple of $-\nabla_\theta \ell_\theta(s, a)$, i.e.\ \emph{parallel} to the SFT descent step: both terms drive $\ell_\theta(s, a)$ down. Crucially, this is not a failure of regularization. It is simply a way to ensure that the policy does not drift too far away from the reference policy.
\end{itemize}

What makes this anchoring safe rather than destructive is that it is (i)~\emph{state-adaptive and silent at the reference}: it activates only as $|r|$ grows, and exactly vanishes at $r = 0$; and (ii)~\emph{sub-dominant to SFT in magnitude}, as made precise in (b) below. So long as the SFT term is weighted strongly enough (which is easily ensured by choosing $\lambda_{\text{SFT}} > \lambda_{\text{PRO}} \cdot \tfrac{\beta}{4}$ in Eq.~\ref{eq:rpro_loss}), the net update on identical-pair samples is dominated by the SFT direction, while the regularizer supplies a bounded correction that keeps the trajectory of $\pi_\theta(a \mid s)$ from straying excessively far from $\pi_{\text{ref}}(a \mid s)$.

\emph{(b)~Bounded magnitude.} Since $|2\sigma(r) - 1| \in [0, 1)$, the regularizer contribution is strictly bounded by a $\tfrac{\beta}{2}$-rescaled version of the SFT gradient norm; in particular it is small near the reference (small $|r|$) and grows only as much as needed to counteract deviation. This is exactly the behavior of a trust-region / KL penalty: it is quiescent inside the trust region and activates smoothly as the policy drifts.

\paragraph{Takeaway: a deliberate design advantage, not a residual artifact.}
Taken together, Steps~1--3 show that for $a^w = a^l$ the FlowPRO objective specializes---without any external gating, indicator function, or sample filtering---to the SFT loss plus a standard trust-region-style regularizer whose magnitude is bounded by $|2\sigma(r) - 1| \in [0, 1)$ and whose direction is, as required of any genuine regularizer, oriented toward $\pi_{\text{ref}}$ in proportion to the deviation: it opposes SFT when $\pi_\theta$ has overshot $\pi_{\text{ref}}$ ($r > 0$) and assists SFT when $\pi_\theta$ still trails $\pi_{\text{ref}}$ ($r < 0$), exactly switching off at $r = 0$. This supports the three properties asserted in the main text (\S\ref{sec:model_design}, Eq.~\ref{eq:gradient_vanishing}): \emph{unified treatment} of contrastive preference pairs and SFT-like samples under a single loss; a \emph{trust-region safeguard} that prevents every gradient step---contrastive or not---from straying too far from $\pi_{\text{ref}}$, which is especially valuable in the offline RL regime where unconstrained SFT updates on scarce corrections would otherwise drift the policy off the data manifold; and \emph{controlled interaction with SFT}, as formalized by Eqs.~\eqref{eq:reg_grad_chain} and \eqref{eq:reg_descent}: the regularizer is one-sided about $\pi_{\text{ref}}$, state-adaptive, and bounded, so SFT demonstrations and positive-trajectory states can be fed through the \emph{same} FlowPRO objective as genuine preference pairs without destabilizing optimization.

\vspace{-1em}
\section{Trajectory Point Distance Metric}
\label{app:distance_metric}

To find the closest point $M'$ on the positive trajectory $\tau^w$ for a given state $M$ on the negative trajectory $\tau^l$, we define a weighted distance metric that jointly considers end-effector position, orientation, and gripper state:
\begin{equation}
d(M, M') = \| \bm{p}_M - \bm{p}_{M'} \|_2 + 0.5 \cdot d_{\text{geo}}(\bm{R}_M, \bm{R}_{M'}) + 0.2 \cdot |g_M - g_{M'}|,
\label{eq:distance_metric}
\end{equation}
where $\bm{p} \in \real^3$ denotes the end-effector position, $d_{\text{geo}}(\bm{R}_M, \bm{R}_{M'}) = \arccos\!\big(\frac{\text{tr}(\bm{R}_M^\top \bm{R}_{M'}) - 1}{2}\big)$ is the geodesic distance between rotation matrices $\bm{R}_M, \bm{R}_{M'} \in SO(3)$, and $g \in [0, 1]$ is the normalized gripper width. The weights 0.5 and 0.2 were determined empirically to balance the contribution of each component.

\vspace{-1em}
\section{Smooth Interpolation Algorithm}
\label{app:interpolation}

Given the current end-effector state at point $M$ on $\tau^l$ and the target action chunk from point $M'$ to $N'$ on $\tau^w$, we construct the synthetic positive action chunk via a two-phase interpolation strategy. Let $n_{\text{tr}} = \max(2, \lfloor \rho H \rfloor)$ with $\rho = 0.7$, and define the \emph{phase-transition point} $J$ as the point reached by advancing $n_{\text{tr}}$ steps from $M'$ along $\tau^w$ (i.e., the $n_{\text{tr}}$-th waypoint of the target chunk). The first phase ($n_{\text{tr}}$ steps) smoothly transitions from $M$ to $J$; the second phase ($H - n_{\text{tr}}$ steps) directly follows the target chunk from $J$ to $N'$. This avoids abrupt jumps while ensuring the synthetic action merges seamlessly into the positive trajectory. The procedure is detailed in Algorithm~\ref{alg:interpolation}.

\begin{algorithm}[h]
\caption{Smooth Interpolation for Synthetic Positive Action Chunk}
\label{alg:interpolation}
\begin{algorithmic}[1]
\REQUIRE Source state $\bm{s}_M = (\bm{p}_0, \bm{R}_0, g_0)$ at point $M$; target chunk $\{(\bm{p}_j, \bm{R}_j, g_j)\}_{j=1}^{H}$ from $\tau^w$; transition ratio $\rho = 0.7$; smoothness $\mu = 0.4$
\ENSURE Interpolated action chunk $\bm{a}^w$ of length $H$
\STATE $n_{\text{tr}} \leftarrow \max(2, \lfloor \rho \cdot H \rfloor)$ \COMMENT{Number of transition steps}
\STATE Let $J$ denote the phase-transition point, i.e., the $n_{\text{tr}}$-th step of the target chunk with state $(\bm{p}_{n_{\text{tr}}}, \bm{R}_{n_{\text{tr}}}, g_{n_{\text{tr}}})$.
\STATE Initialize $\bm{a}^w \leftarrow$ copy of target chunk
\FOR{each arm}
    \STATE \textbf{// Phase 1: Smooth transition from $M$ to $J$ (steps $1, \ldots, n_{\text{tr}}$)}
    \STATE \textbf{Position (cubic Bezier):}
    \STATE \quad $P_0 \leftarrow \bm{p}_0$, \quad $P_3 \leftarrow \bm{p}_{n_{\text{tr}}}$
    \STATE \quad $P_1 \leftarrow (P_0 + P_3) / 2$ \COMMENT{No source tangent to avoid bending}
    \STATE \quad Estimate target tangent $\hat{\bm{d}}$ at $J$ from $\bm{p}_{n_{\text{tr}}-1}, \bm{p}_{n_{\text{tr}}+1}$
    \STATE \quad $P_2 \leftarrow P_3 - \mu \cdot \|P_3 - P_0\| \cdot \hat{\bm{d}}$ \COMMENT{End tangent for smooth arrival}
    \FOR{$i = 1$ to $n_{\text{tr}}$}
        \STATE $t \leftarrow (i-1)/(n_{\text{tr}}-1)$
        \STATE $\bm{a}^w_i[\text{pos}] \leftarrow (1{-}t)^3 P_0 + 3(1{-}t)^2 t \, P_1 + 3(1{-}t) t^2 P_2 + t^3 P_3$
    \ENDFOR
    \STATE \textbf{Orientation (Slerp):}
    \FOR{$i = 1$ to $n_{\text{tr}}$}
        \STATE $t \leftarrow (i-1)/(n_{\text{tr}}-1)$
        \STATE $\bm{a}^w_i[\text{rot}] \leftarrow \text{Slerp}(\bm{R}_0, \bm{R}_{n_{\text{tr}}}, t)$
    \ENDFOR
    \STATE \textbf{Gripper (linear):}
    \FOR{$i = 1$ to $n_{\text{tr}}$}
        \STATE $t \leftarrow (i-1)/(n_{\text{tr}}-1)$
        \STATE $\bm{a}^w_i[\text{grip}] \leftarrow (1-t) \cdot g_0 + t \cdot g_{n_{\text{tr}}}$
    \ENDFOR
    \STATE \textbf{// Phase 2: Follow target from $J$ onward (steps $n_{\text{tr}}{+}1, \ldots, H$)}
    \STATE Steps $n_{\text{tr}}{+}1$ through $H$ retain the original target chunk values.
\ENDFOR
\RETURN $\bm{a}^w$
\end{algorithmic}
\end{algorithm}

The cubic Bezier curve uses only the \emph{terminal} tangent $P_2$ (derived from the target trajectory) while setting $P_1$ to the midpoint of $P_0$ and $P_3$. This design choice eliminates the influence of the source trajectory's motion direction at $M$, which may point toward the erroneous region, and ensures that the interpolated path smoothly converges to the positive trajectory without undesired bending.

\section{Physical Plausibility of Smooth Interpolation}
\label{app:physical_plausibility}

\begin{figure}[h]
    \centering
    \includegraphics[width=0.3\linewidth]{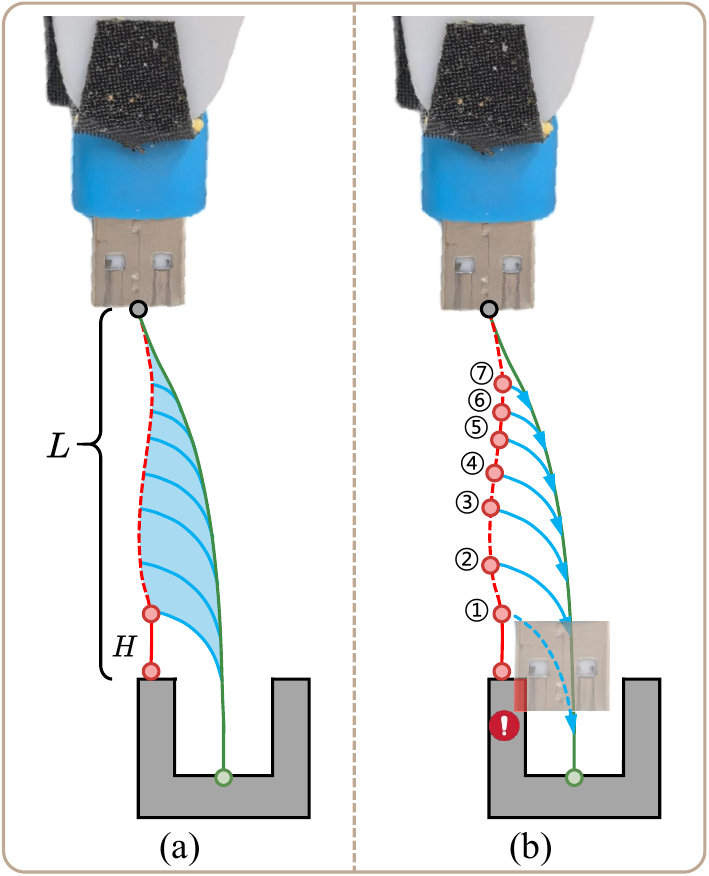}
    \caption{\textbf{Why Smooth Interpolation stays physically plausible}, illustrated on the \textsc{USB} task. In both panels: \textbf{red dashed} = executed negative rollout $\tau^l$ (length $L$, ending at the socket rim); \textbf{green solid} = successful teleoperation $\tau^w$ (ending inside the socket); \textbf{blue arc} = one synthetic positive chunk of length $H$, a cubic B\'ezier bridge from $M\!\in\!\tau^l$ to $J\!\in\!\tau^w$ followed by direct tracking of $\tau^w$ up to $N'$. \textbf{(a) Generic case.} Every synthetic chunk lies in the blue lens-shaped region between $\tau^l$ and $\tau^w$, safely away from the socket wall. \textbf{(b) Worst case.} Even if state \textcircled{1} (next to the socket rim) gave rise to a wall-clipping chunk (dashed blue arc), the policy never reaches \textcircled{1} at deployment: at every preceding state \textcircled{2}--\textcircled{7} along $\tau^l$, the per-state positive action $a^w$ (solid blue arcs) already steers it onto $\tau^w$.}
    \label{fig:safe_interpolation}
    \vspace{-2em}
\end{figure}

A natural concern about Smooth Interpolation (Case~1 in Sec.~\ref{sec:data_processing}) is whether the cubic-B\'ezier bridge from $M\!\in\!\tau^l$ to $J\!\in\!\tau^w$ can synthesize \emph{physically infeasible} motions---for example, in the \textsc{USB} task, a synthetic chunk that geometrically intersects with the USB socket wall. Fig.~\ref{fig:safe_interpolation} illustrates the geometric picture on the \textsc{USB} task; we argue below that this risk is suppressed by three complementary properties of our data-collection and interpolation pipeline, and that any residual risk is irrelevant at deployment time.

\paragraph{(i) High sampling rate bounds the per-chunk geometric displacement.}
All teleoperation data are recorded at $50$\,Hz, and Smooth Interpolation operates on a single action chunk of length $H$ (a fraction of a second in our experiments). The synthetic bridge therefore spans only the distance the end-effector travels in $\rho H \!=\! 0.7H$ steps---typically a few millimeters to a centimeter in our tasks. By construction, the cubic B\'ezier curve lies in the convex hull of its four control points $\{P_0,\ldots,P_3\}$, all of which are anchored to $M\!\in\!\tau^l$, $J\!\in\!\tau^w$, or their midpoint and terminal tangent. The synthetic chunk is therefore confined to a thin lens-shaped neighborhood between $\tau^l$ and $\tau^w$---the blue region in Fig.~\ref{fig:safe_interpolation}(a)---and large excursions into obstacles are geometrically ruled out.

\paragraph{(ii) The sampling range of $M$ excludes the dangerous tail of $\tau^l$.}
The only situation in which $\tau^l$ comes close to an obstacle is at its very end: the rollback signal is triggered \emph{precisely} when the operator detects an imminent collision (e.g., the USB head touching the socket rim), so collisions, if they occur at all on $\tau^l$, are confined to the last few steps before the rollback timestamp. To eliminate this case at the source, we restrict the source point $M$ to be drawn from the prefix $[0,\,L-H]$ of $\tau^l$, where $L$ is the length of $\tau^l$ and $H$ is the chunk length (cf.\ the labels $L$ and $H$ on the $\tau^l$ side of Fig.~\ref{fig:safe_interpolation}(a)). This guarantees (a)~that the entire negative chunk $a^l$ used in the loss lies strictly inside $\tau^l$ (rather than past its end), and (b)~that $M$ is at least one chunk away from the potentially-contact-rich terminal segment. The dangerous tail of $\tau^l$ is therefore never used as the starting point of an interpolation bridge.

\paragraph{(iii) RPRO ensures the residual-risk states are never visited at deployment.}
Even in the worst case where $M$ is allowed to land arbitrarily close to the socket rim, the resulting offending chunk is harmless at deployment time, because the policy never reaches the dangerous state. Concretely, consider state~\textcircled{1} in Fig.~\ref{fig:safe_interpolation}(b), which sits right next to the obstacle: the synthetic chunk fired from \textcircled{1} (blue dashed line) would clip into the socket wall. 
However, at every preceding state \textcircled{2}--\textcircled{7} along $\tau^l$ before \textcircled{1}, the per-state positive action $a^w$ (blue solid arcs) already steers the policy onto $\tau^w$, so \textcircled{1} is \emph{never reached} at deployment. Empirically, the per-iteration success curves in Fig.~\ref{fig:baseline_curves} confirm that the policy no longer drifts toward the $\tau^l$ tail, and across all $K{=}3$ rounds and four tasks in Table~\ref{tab:main_results}, we observed no rollout failure attributable to interpolation artifacts.

Taken together, (i)~the small per-chunk displacement at $50$\,Hz keeps every synthetic bridge inside the lens-shaped safe region of Fig.~\ref{fig:safe_interpolation}(a), (ii)~the $[0,\,L-H]$ sampling rule for $M$ removes the dangerous tail of $\tau^l$ from interpolation at the source, and (iii)~RPRO training renders any residual-risk states (such as state~\textcircled{1} in Fig.~\ref{fig:safe_interpolation}(b)) unreachable at deployment time.

\section{Iterative Training Procedure}
\label{app:training}

The full iterative training procedure of FlowPRO is given in Algorithm~\ref{alg:training}.

\begin{algorithm}[H]
\caption{Iterative Training of FlowPRO}
\label{alg:training}
\begin{algorithmic}[1]
\REQUIRE SFT dataset $\dataset_{\text{SFT}}$; reference policy $\pi_{\text{ref}}$; number of iterations $K$; loss coefficients $\lambda_{\text{PRO}}, \lambda_{\text{SFT}}$; reward-proxy temperature $\beta$; action-chunk horizon $H$ (see App.~\ref{app:hardware})
\ENSURE Fine-tuned policy $\pi_\theta$
\STATE \textbf{Stage 1 (Initial SFT):}
\STATE Train $\pi_\theta$ on $\dataset_{\text{SFT}}$ with $\loss_{\text{SFT}}$
\STATE \textbf{Stage 2 (Iterative Offline-RL Fine-Tuning):}
\FOR{$k = 1$ to $K$}
    \STATE $\pi_{\text{ref}} \leftarrow \pi_\theta$ \COMMENT{Reference = policy from previous iteration}
    \STATE \textbf{// Data Collection}
    \STATE Roll out $\pi_\theta$; human intervenes via teleoperation
    \STATE Obtain trajectory pairs $(\tau^w_k, \tau^l_k)$
    \STATE Construct $\dataset_{\text{pref}}^k$ via interpolation
    \STATE \textbf{// Training}
    \FOR{each training step}
        \STATE \textbf{// Batch Composition}
        \IF{$k = 1$}
            \STATE Sample mini-batch $\mathcal{B} = \{(s, a^w, a^l)\}$ with 80\% from $\dataset_{\text{pref}}^k$ and 20\% from $\dataset_{\text{SFT}}$
        \ELSE
            \STATE Sample mini-batch $\mathcal{B} = \{(s, a^w, a^l)\}$ with 70\% from $\dataset_{\text{pref}}^k$, 15\% from $\dataset_{\text{pref}}^{<k}$, and 15\% from $\dataset_{\text{SFT}}$
        \ENDIF
        \STATE \textbf{// Update}
        \STATE $\loss \leftarrow \lambda_{\text{PRO}} \cdot \loss_{\text{PRO}}(\theta; \mathcal{B}) + \lambda_{\text{SFT}} \cdot \loss_{\text{SFT}}(\theta; \mathcal{B})$
        \STATE Update $\theta$ via $\nabla_\theta \loss$
    \ENDFOR
\ENDFOR
\RETURN $\pi_\theta$
\end{algorithmic}
\end{algorithm}

\section{Hardware Platform and Training Setup}
\label{app:hardware}

\subsection{Robot Platform}
\label{app:hardware_robot}

\begin{figure}[h]
    \centering
    \includegraphics[width=0.8\linewidth]{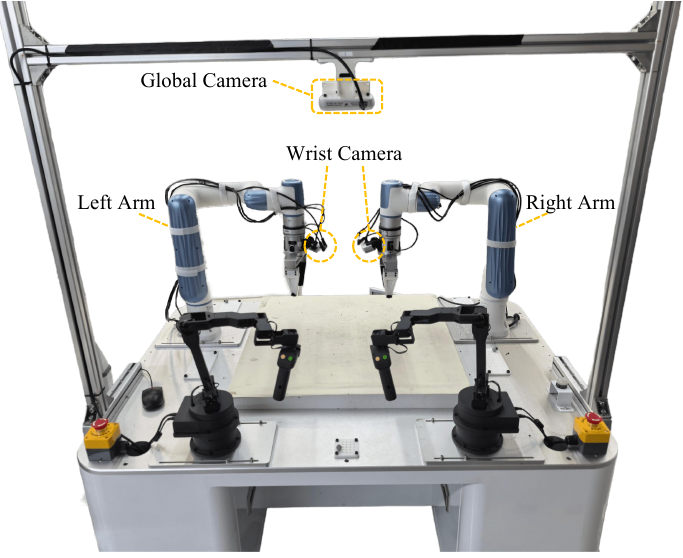}
    \caption{\textbf{Hardware platform used for all real-robot experiments.} The Dobot XTrainer bimanual setup includes two 6-DoF arms with parallel-jaw grippers, a top-down global RGB camera, two wrist-mounted RGB cameras (one per arm).}
    \label{fig:experiment_set_up}
\end{figure}

All real-robot experiments are conducted on a \textbf{Dobot XTrainer} bimanual teleoperation platform. The platform is equipped with:
\begin{itemize}[leftmargin=1.2em]
    \item Two 6-DoF robot arms with parallel-jaw grippers (left and right);
    \item Three RGB cameras: one top-down global camera, and one wrist camera on each arm;
    \item A Meta Quest~3 headset used by the human operator for both VR teleoperation during SFT/correction data collection and for the intervention-and-rollback interface described in \S\ref{sec:data_collection}.
\end{itemize}
A photograph of the platform with all of the above components labeled in situ is shown in Fig.~\ref{fig:experiment_set_up}.

\subsection{Training Setup}
\label{app:training_setup}

This subsection details the compute resources, data scale, and all training hyper-parameters used to produce every model evaluated in the main paper. Both the SFT base policy and all RPRO iterations are trained on the same compute node and share the same image and proprioception pipeline.

\paragraph{Compute and precision.}
All training runs are executed on a single node with $4\times$NVIDIA~H20 GPUs (96~GB each) under \texttt{bfloat16} mixed precision, with a peak per-GPU memory footprint of approximately~90~GB. Inference of the flow-matching action expert uses $10$ denoising steps per chunk.

\paragraph{Action representation.}
The policy outputs an action chunk of length $H = 50$ steps, which corresponds to $1.0$~s of motion at the $50$~Hz control rate stated in \S\ref{sec:method} (item~(i) of the physical-plausibility analysis).

\paragraph{Stage~1: SFT base policy.}
The base policy is trained by flow-matching regression on a per-task demonstration dataset. The numbers of expert episodes per task are summarized in Table~\ref{tab:sft_demos}. Training is run for $60\,000$ optimizer steps with a per-GPU batch size of $16$ (global batch size $64$ across the $4$ GPUs). The optimizer is AdamW with an initial learning rate of $2.5\times 10^{-5}$, a linear warmup over the first $1{,}000$ steps, followed by a cosine decay over the next $30{,}000$ steps to a floor learning rate of $2.5\times 10^{-6}$, which is then held constant for the remaining steps.

\begin{table}[h]
\centering
\small
\caption{\textbf{Number of expert demonstration episodes used for Stage~1 (SFT) per task.}}
\label{tab:sft_demos}
\vspace{6pt}
\begin{tabular}{lcccc}
\toprule
Task & \textsc{Pack} & \textsc{Cap} & \textsc{USB} & \textsc{Case} \\
\midrule
\#~SFT episodes & 773 & 298 & 389 & 300 \\
\bottomrule
\end{tabular}
\end{table}

\paragraph{Stage~2: RPRO iterative fine-tuning.}
At each RPRO iteration $k \in \{1, 2, 3\}$, the policy is fine-tuned for $25\,000$ optimizer steps (totalling $75\,000$ steps across all three rounds) with a \emph{global} batch size of $20$ (i.e., $5$ samples per GPU). The optimizer is AdamW with an initial learning rate of $1\times 10^{-5}$, a linear warmup over $1{,}000$ steps, followed by a cosine decay over the next $15{,}000$ steps to a floor of $2.5\times 10^{-6}$. The batch composition follows Algorithm~\ref{alg:training}: at $k=1$, mini-batches mix the current preference set $\dataset_{\text{pref}}^{1}$ and the SFT dataset in an $80/20$ ratio; at $k \geq 2$, they mix $\dataset_{\text{pref}}^{k}$, the union of past preference sets $\dataset_{\text{pref}}^{<k}$, and $\dataset_{\text{SFT}}$ in a $70/15/15$ ratio.

The number of preference pairs $(\tau^w, \tau^l)$ collected through the teleoperated intervention pipeline at each round and for each task is reported in Table~\ref{tab:pref_pairs}. Each pair is subsequently expanded into per-state $(s, a^w, a^l)$ tuples via the smooth interpolation procedure of \S\ref{sec:data_processing} before being mixed into the training batches.

\begin{table}[h]
\centering
\small
\caption{\textbf{Number of preference pairs $(\tau^w_k, \tau^l_k)$ collected at each RPRO round for each task.}}
\label{tab:pref_pairs}
\vspace{6pt}
\begin{tabular}{lcccc}
\toprule
Round & \textsc{Pack} & \textsc{Cap} & \textsc{USB} & \textsc{Case} \\
\midrule
$k = 1$ & 40 & 89 & 30 & 63 \\
$k = 2$ & 30 & 42 & 30 & 43 \\
$k = 3$ & 30 & 30 & 30 & 30 \\
\bottomrule
\end{tabular}
\end{table}

\paragraph{Loss coefficients and reward-proxy temperature.}
The composite objective of Algorithm~\ref{alg:training} (inner-loop update) and Eq.~\ref{eq:rpro_loss} uses $\lambda_{\text{PRO}} = 1/3$ and $\lambda_{\text{SFT}} = 1$, applied to every batch throughout all three RPRO rounds. The reward-proxy temperature in Eq.~\ref{eq:reward_proxy} and Eq.~\ref{eq:pro_loss} is set to $\beta = 3.5$. These three scalars are the only RPRO-specific hyper-parameters introduced on top of the base flow-matching pipeline and are shared across all four tasks.

\paragraph{Reproducibility.}
Each cell of the main results table is averaged over $3$ random seeds (data shuffling and parameter initialization), and each seed is evaluated for $100$ rollouts under both the ID and OOD protocols described in \S\ref{sec:tasks}. All other hyper-parameters not listed above are inherited unchanged from the base flow-matching policy.

\section{Task Details}
\label{app:tasks}

\begin{figure*}[t]
    \centering
    \includegraphics[width=\textwidth]{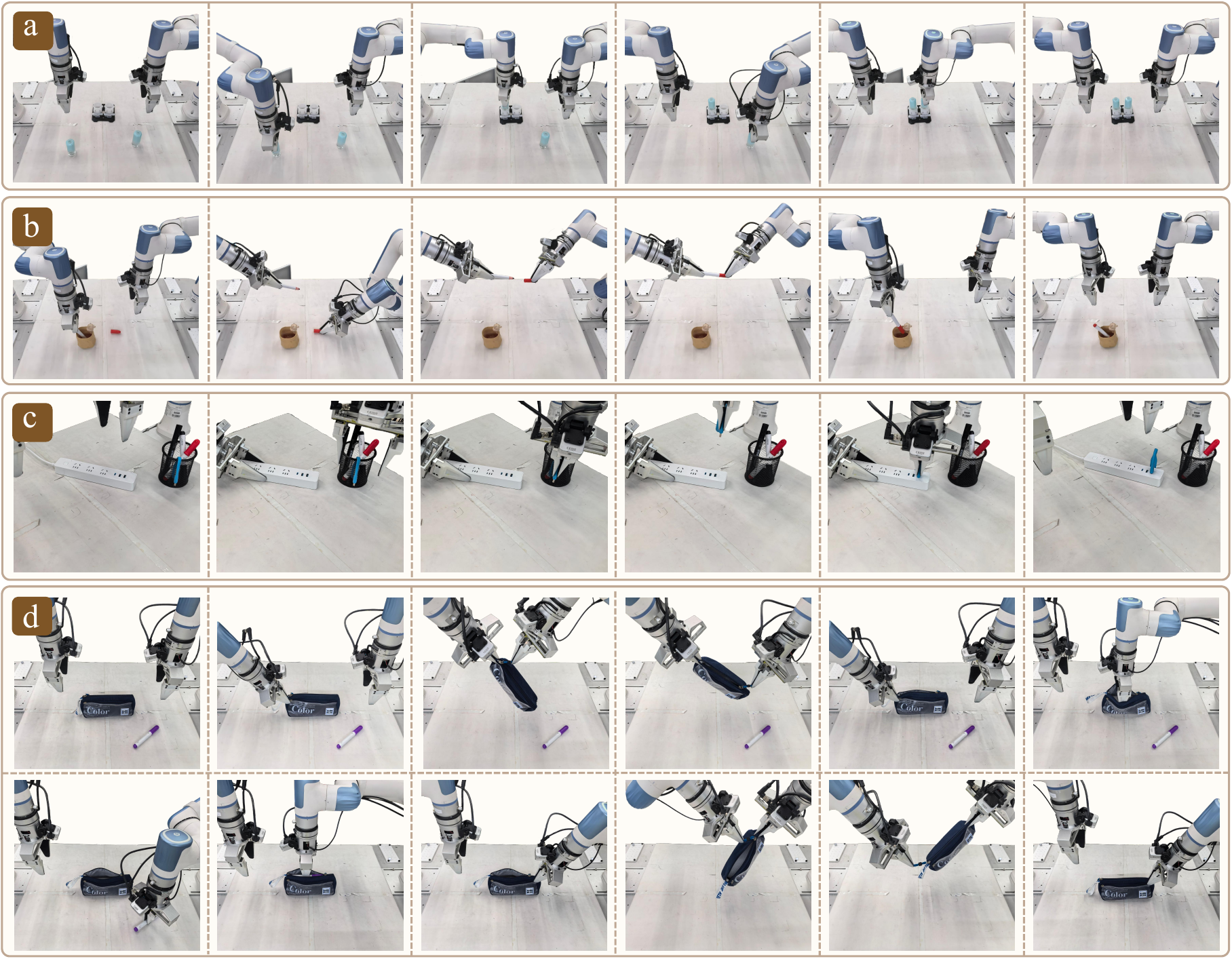}
    \caption{\textbf{Complete per-stage pipeline for the four real-robot tasks.} Extended version of Fig.~\ref{fig:experiment_task} in the main text, showing every sub-stage of \textsc{Pack}, \textsc{Cap}, \textsc{USB}, and \textsc{Case}.}
    \label{fig:complete_experiment_task}
\end{figure*}

This appendix expands the brief task descriptions in \S\ref{sec:tasks} with the detailed setup, per-stage sub-goals, and the dominant failure modes of the SFT policy that FlowPRO is designed to repair. The complete per-stage pipeline for all four tasks is illustrated in Fig.~\ref{fig:complete_experiment_task}.

\paragraph{Cosmetic Packaging (\textsc{Pack}).}
Two cosmetic containers are placed at fixed marker regions on either side of a bimanual packaging holder, with small per-episode randomization of their in-plane position. The left arm grasps the left container and inserts it into the left slot of the holder, then the right arm grasps the right container and inserts it into the right slot. The dominant difficulty lies in the \emph{insertion} phase, which requires sub-centimeter positional accuracy and tight alignment of the container axis with the slot axis. Near the end of the trajectory, two failure modes dominate: (i) the container collides with the rim of the slot and gets pushed aside, and (ii) the container is inserted at a slight tilt and wedges before reaching the bottom. Both failures occur in the final fraction of the trajectory and are the hardest to recover from via SFT alone, since SFT receives no learning signal from negative outcomes.

\paragraph{Pen-Cap Assembly (\textsc{Cap}).}
The left arm picks an uncapped pen from a pen holder, then the right arm picks a pen cap from the table; the two arms then meet in mid-air and the right arm presses the cap onto the pen tip. The pens and the cap are placed at fixed marker regions with randomization. Three distinct difficulties make this task a challenging stress test for a VLA policy:
\begin{itemize}[leftmargin=1.2em]
    \item \emph{Left-arm grasp of the upright pen.} Several pens stand upright in the holder with only a short segment exposed above the rim, so the left arm has a very limited graspable region. The SFT policy frequently misses the pen entirely or only catches it near the edge of the fingers, leading to slippage during lift-off.
    \item \emph{Right-arm grasp of the flat cap.} The pen cap lies flat on the table and is very thin, so the right gripper must align precisely with its low profile. Small vertical or yaw errors cause the fingers to either miss the cap completely or only pinch its edge, which makes the cap tilt or drop shortly after being lifted.
    \item \emph{In-air bimanual assembly.} Even when both grasps succeed, the two arms must meet at a free-floating rendezvous point and align the cap-to-pen axis with sub-centimeter accuracy. Because both arms are moving and neither object is supported by the table, small pose errors on \emph{either} arm are amplified at the contact point, and the assembly is highly sensitive to residual errors inherited from the grasping phase.
\end{itemize}
These three difficulties appear at different temporal segments of the rollout (early, early, late) and involve different arms, which makes the task a good testbed for methods that need to correct localized failure modes without disturbing the already-working behavior elsewhere in the trajectory.

\paragraph{USB Insertion (\textsc{USB}).}
The left arm grasps a power strip placed at a marker region on the table and lifts it to a stable holding pose, while in parallel the right arm picks a small USB lamp standing in a pen holder; the right arm then inserts the lamp's USB plug into one of the strip's sockets. Object positions are fixed at marker regions with per-episode randomization of in-plane position and yaw. The dominant difficulty lies in the final \emph{plug-in-socket insertion}: USB connectors require sub-millimeter accuracy in both position and orientation, and the socket aperture leaves essentially no clearance, so even a small residual error in either arm prevents the plug from entering. Two failure modes dominate near the end of the trajectory: (i) the plug edge collides with the socket bezel and is pushed off-axis, and (ii) the plug enters at a slight tilt and stalls before fully seating. Because the left arm is also actively holding the strip rather than resting on a fixture, both arms contribute pose error simultaneously, making this task a stringent test of \emph{coordinated high-precision insertion}.

\paragraph{Pencil-Case Packing (\textsc{Case}).}
This task is the most long-horizon in our suite, comprising six chained sub-stages executed by two arms in close coordination:
\begin{enumerate}[leftmargin=1.4em,itemsep=1pt,topsep=1pt]
    \item the left arm picks up the closed soft-fabric pencil case from a marker region on the table;
    \item the right arm grasps the zipper slider, and the two arms cooperate to \emph{unzip} the case;
    \item the left arm places the now-open case at the table center;
    \item the right arm reaches inside the case and holds it open from within;
    \item the right arm picks a pen lying on the table and drops it into the case;
    \item the right arm grasps the pencil case again and the two arms cooperate to \emph{zip the case closed} and return it to the table center.
\end{enumerate}
Object positions are fixed at marker regions with per-episode randomization. Three properties together make this task uniquely challenging: \emph{(a) length and chaining}---a single early-stage error (e.g., a slipped grasp at stage~1 or an off-axis pull at stage~2) propagates and contaminates all subsequent stages, so the SFT policy's success rate compounds multiplicatively across stages; \emph{(b) high-precision zipping}---both unzip (stage~2) and zip (stage~6) require the right gripper to align with and pull the small slider along a curved fabric track, which is geometrically as demanding as the insertion tasks above but must be executed twice in a single rollout; \emph{(c) deformability}---the fabric case continuously changes shape under contact, so the visual configuration the policy observes at each stage drifts away from the SFT distribution depending on prior interactions, and conventional SFT supervision provides little signal for recovering from such state-dependent deformation. Failures observed under SFT span all stages: zipper grip slip, asymmetric two-arm pulling that jams the zipper, the case folding while the pen is being inserted.

\section{Baseline Methods}
\label{app:baseline_details}

This appendix expands the brief baseline summary in \S\ref{sec:baselines} with the full methodological description of each comparator. The corresponding loss formulations in the flow-matching setting are deferred to Appendix~\ref{app:ablation_losses}.

\paragraph{DAgger~\citep{ross2011dagger}.}
\emph{Positive-only} interactive-imitation baseline via \emph{dataset aggregation}. At each RPRO round, expert corrections collected on failed states through our teleoperated intervention pipeline (\S\ref{sec:data_collection}) are merged directly into the SFT dataset, and the policy is re-trained from scratch by behavior cloning on the aggregated set. This setting captures the dominant ``correct-then-retrain'' paradigm used in prior interactive imitation work.

\paragraph{DAgger-Buffered.}
\emph{Positive-only} baseline via \emph{controlled batch mixing}---an internal control we introduce to isolate the contribution of the preference loss from that of our batch-composition schedule. New expert corrections are kept in a separate buffer rather than being merged into the SFT dataset, and each training batch combines the SFT dataset and the buffered data using the \emph{same} batch-composition schedule as FlowPRO. Comparing DAgger-Buffered against DAgger isolates the contribution of batch composition, while comparing it against FlowPRO isolates the contribution of the preference-based loss itself.

\paragraph{PI0.6*~\citep{pistar06_2025}.}
\emph{Positive-and-negative} baseline via \emph{advantage conditioning}. The flow-matching expert is reconditioned on a binary improvement indicator $I_t \in \{0, 1\}$ derived from our preference labels: states from positive trajectories $\tau^w$ are labelled $I_t=1$ and states from negative trajectories $\tau^l$ are labelled $I_t=0$. At inference, classifier-free guidance is applied with respect to $I_t$ to steer generation toward improved actions. This comparator isolates the difference between a \emph{conditioning}-based and a \emph{contrastive-loss}-based use of identical preference data.

\paragraph{TPO~\citep{zhang2025grape}.}
\emph{Positive-and-negative} baseline via \emph{trajectory-wise preference optimization}. Our state-wise reward proxy (Eq.~\ref{eq:reward_proxy}) is summed over each trajectory to form a trajectory-level reward $r_\theta(\tau) = \sum_t r_\theta(s_t, a_t)$, which is then plugged into the FlowPRO objective over $(\tau^w, \tau^l)$ pairs in place of state-level rewards. This comparator isolates per-state vs.~trajectory-level preference construction while keeping the loss family fixed.

\begin{figure}[h]
    \centering
    \includegraphics[width=\textwidth]{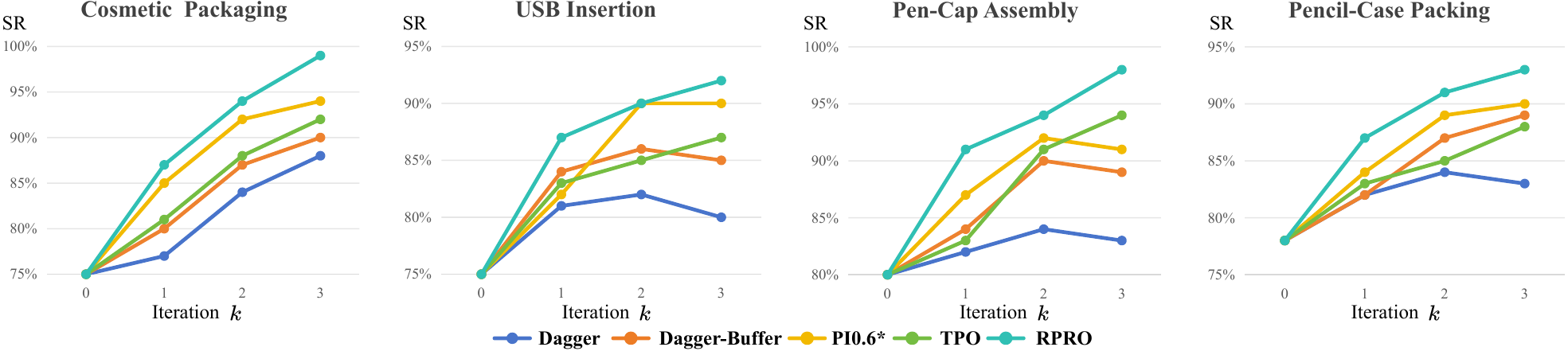}
    \caption{\textbf{Per-iteration success rates (SR) on the four real-robot tasks, with \textsc{PI0} as the base policy.} Iteration~$0$ corresponds to the shared SFT checkpoint. Companion to Fig.~\ref{fig:baseline_curves} in the main text, which reports the analogous curves with \textsc{PI0.5} as the base policy.}
    \label{fig:baseline_curves_pi0}
\end{figure}

\paragraph{Per-iteration baseline curves on \textsc{PI0}.}
For completeness, Fig.~\ref{fig:baseline_curves_pi0} reports the per-iteration success rates of all comparators when the base policy is \textsc{PI0} instead of \textsc{PI0.5}, matching the \textsc{PI0.5} curves shown in Fig.~\ref{fig:baseline_curves} of the main text.

\section{Loss Formulations for Baselines and Ablations}
\label{app:ablation_losses}

For completeness, we list here the exact loss expressions adopted for each ablation variant in the flow-matching setting. Throughout this appendix, $\ell_\theta(s, a) = \expect_{t, \epsilon} [ \| v_\theta(a_t, t | s) - u(a_t | a) \|^2 ]$ denotes the per-sample flow-matching loss and $r_\theta(s, a) = \tfrac{\beta}{2}(\ell_{\text{ref}}(s, a) - \ell_\theta(s, a))$ is the implicit reward proxy (Eq.~\ref{eq:reward_proxy}).

\paragraph{DAgger.}
A \emph{positive-only} interactive imitation baseline that adapts the classical DAgger~\citep{ross2011dagger} pipeline to our preference-collection protocol. At round $k$, the policy $\pi_{\theta_{k-1}}$ is rolled out on the target tasks, an expert relabels the failed states via the teleoperated intervention pipeline (\S\ref{sec:data_collection}), and the resulting correction set $\dataset_k$ (containing only the \emph{preferred} actions $a^w$) is \emph{aggregated} directly into the running SFT dataset:
\begin{equation}
\dataset_{\text{SFT}}^{(k)} = \dataset_{\text{SFT}}^{(k-1)} \cup \{(s, a^w) : (s, a^w, a^l) \in \dataset_k\}, \quad \dataset_{\text{SFT}}^{(0)} = \dataset_{\text{SFT}}.
\end{equation}
The policy at round $k$ is then re-trained by behavior cloning on the aggregated set:
\begin{equation}
\loss_{\text{DAgger}}^{(k)}(\theta) = \expect_{(s, a) \sim \dataset_{\text{SFT}}^{(k)}} \big[ \ell_\theta(s, a) \big].
\end{equation}
Negative trajectories $\tau^l$ are discarded; DAgger therefore consumes the same expert annotation budget as FlowPRO but only exploits the positive half of each preference pair, and uniform aggregation gives no control over the SFT-to-correction sample ratio in each batch.

\paragraph{DAgger-Buffered.}
A \emph{positive-only} ablation of DAgger that isolates the effect of \emph{controlled batch mixing} from the underlying loss family. Instead of merging expert corrections into the SFT pool, we keep three separate buffers in exact correspondence with FlowPRO: the round-$k$ correction set $\dataset_{\text{corr}}^k = \{(s, a^w) : (s, a^w, a^l) \in \dataset_k\}$, the historical pool $\dataset_{\text{corr}}^{<k} \triangleq \bigcup_{j<k}\dataset_{\text{corr}}^j$, and $\dataset_{\text{SFT}}$. Let $\mathcal{P}_{\text{mix}}^{(k)}$ denote the mixture sampling distribution over these three buffers under the \emph{same fixed proportions as FlowPRO} (Sec.~\ref{sec:data_processing}; $80\%/20\%$ for $k{=}1$ and $70\%/15\%/15\%$ for $k{\geq}2$). The policy is trained by behavior cloning under this mixture:
\begin{equation}
\loss_{\text{DAgger-Buf}}^{(k)}(\theta) = \expect_{(s, a^w) \sim \mathcal{P}_{\text{mix}}^{(k)}} \big[ \ell_\theta(s, a^w) \big].
\end{equation}
Compared to vanilla DAgger, only the data buffering and mixing schedule changes; the loss family (positive-only behavior cloning), the expert annotations used, and the optimization protocol are identical. This baseline therefore quantifies how much of the benefit of our pipeline comes from \emph{batch composition} alone, independent of the contrastive preference objective.

\paragraph{TPO (trajectory-wise).}
Following the TPO objective of GRAPE~\citep{zhang2025grape}, this baseline applies a trajectory-wise preference loss to the \emph{whole-trajectory log-likelihood ratio}. Under the MDP assumption $\pi(\zeta) = \prod_{t=1}^{T} \pi(a_t \mid o_t)$, so $\log\frac{\pi_\theta(\zeta)}{\pi_{\text{ref}}(\zeta)} = \sum_{t=1}^{T}\log\frac{\pi_\theta(a_t \mid o_t)}{\pi_{\text{ref}}(a_t \mid o_t)}$. In the flow-matching setting, we approximate the per-step log-ratio using the flow-matching loss surrogate and define the trajectory-level reward proxy
\begin{equation}
r_\theta(\tau) = \sum_{(s_t, a_t) \in \tau} r_\theta(s_t, a_t) = \tfrac{\beta}{2} \sum_{(s_t, a_t) \in \tau} \big(\ell_{\text{ref}}(s_t, a_t) - \ell_\theta(s_t, a_t)\big).
\end{equation}
This trajectory-level proxy is plugged into the RPRO objective (Eq.~\ref{eq:rpro_loss}) rather than the DPO objective used in the original GRAPE formulation, so that TPO and FlowPRO share the same preference loss family and differ only in the granularity of the reward proxy (trajectory-level vs.\ step-level), combined with the SFT term on the preferred trajectory:
\begin{equation}
\loss_{\text{TPO}}(\theta) = \loss_{\text{PRO}}\big(r_\theta(\tau^w), r_\theta(\tau^l)\big) + \lambda_{\text{SFT}} \cdot \loss_{\text{SFT}}(\theta).
\end{equation}
The per-step reward proxy $r_\theta(s_t, a_t)$ is shared with RPRO (Eq.~\ref{eq:reward_proxy}); the only difference is whether the optimizer operates on a single-state reward gap or on the trajectory-summed gap.

\paragraph{PI0.6*.}
Following RECAP~\citep{pistar06_2025}, PI0.6* adds \emph{advantage conditioning} to the flow-matching action expert rather than modifying its loss. The full RECAP pipeline trains a distributional value function $p_\phi(V\mid o_t, \ell)$ and computes an $N$-step advantage $A^\pi(o_t, a_t)$ that is binarized into an improvement indicator $I_t = \mathbb{1}[A^\pi > \epsilon_\ell]$. In our setting, the teleoperated intervention-and-rollback pipeline (\S\ref{sec:data_collection}) already provides trajectory-level positive/negative labels that directly serve as $I_t$:
\begin{equation}
I_t = \begin{cases}
\texttt{True}, & \text{if the state is on the positive trajectory } \tau^w \text{ (or is an SFT expert sample)}, \\
\texttt{False}, & \text{if the state is on the negative trajectory } \tau^l.
\end{cases}
\end{equation}
We therefore bypass RECAP's value-function training and advantage-estimation machinery, and reuse only the policy-side optimization. The flow-matching action expert is reconditioned on $I_t$ as an additional input token, and trained with a standard conditional flow-matching loss plus a 30\% indicator dropout (to enable classifier-free-guidance sampling at inference):
\begin{equation}
\loss_{\text{RECAP}}(\theta) = \expect_{(o, a, I_t)} \big[ \| v_\theta(a_\eta, \eta \mid o, \ell, I_t) - (a - \omega) \|^2 \big], \quad \text{with } I_t \text{ masked w.p.~} 0.3,
\end{equation}
where $a_\eta = \eta a + (1-\eta)\omega$ with $\omega\sim\mathcal{N}(0, I)$ is the flow-matching interpolant and $\eta\in[0,1]$ is the flow time index. At inference time we fix $I_t=\texttt{True}$ and optionally apply classifier-free guidance between the conditional and unconditional velocity fields to further amplify the ``improvement'' direction. This formulation isolates the contribution of \emph{advantage conditioning as a policy-improvement mechanism}, holding the preference dataset and label assignment identical to FlowPRO.

\paragraph{SFT.}
Pure flow-matching regression on preferred actions $a^w$ only:
\begin{equation}
\loss_{\text{SFT}}(\theta) = \expect_{(s, a^w) \sim \dataset} \big[ \ell_\theta(s, a^w) \big].
\end{equation}

\paragraph{DPO (in flow matching).}
Standard pairwise DPO loss with the implicit reward proxy:
\begin{equation}
\loss_{\text{DPO}}(\theta) = -\expect_{(s, a^w, a^l) \sim \dataset} \Big[ \log \sigma\!\big( r_\theta(s, a^w) - r_\theta(s, a^l) \big) \Big].
\end{equation}
Only the contrastive term is present; there is no proximal regularizer and no SFT term.

\paragraph{DPO + SFT.}
DPO loss augmented with an SFT regression on the \emph{chosen} action only:
\begin{equation}
\loss_{\text{DPO+SFT}}(\theta) = \lambda_{\text{PRO}} \cdot \loss_{\text{DPO}}(\theta) + \lambda_{\text{SFT}} \cdot \expect_{(s, a^w) \sim \dataset} \big[ \ell_\theta(s, a^w) \big].
\end{equation}
Note that $\loss_{\text{DPO}}$ here is exactly the contrastive optimizer $\loss_{\text{con}}$ inside $\loss_{\text{PRO}}$ (Eq.~\ref{eq:pro_loss}); DPO+SFT therefore corresponds to RPRO with the proximal regularizer $\loss_{\text{reg}}$ removed, and uses the same coefficients $\lambda_{\text{PRO}} = 1/3$, $\lambda_{\text{SFT}} = 1$ as RPRO.

\paragraph{PRO (no SFT).}
The contrastive-plus-regularizer component $\loss_{\text{PRO}}$ (Eq.~\ref{eq:pro_loss}) used \emph{alone}, i.e., the full RPRO loss (Eq.~\ref{eq:rpro_loss}) with $\lambda_{\text{SFT}}=0$:
\begin{equation}
\begin{split}
\loss_{\text{PRO}}(\theta) = -\expect_{(s, a^w, a^l)} \Big[ &\log \sigma(r_\theta(s, a^w) - r_\theta(s, a^l)) \\
&+ \sum_{a \in \{a^w, a^l\}} \big( \tfrac{1}{2}\log\sigma(r_\theta(s, a)) + \tfrac{1}{2}\log\sigma(-r_\theta(s, a)) \big) \Big].
\end{split}
\end{equation}

\paragraph{RPRO (full).}
The full RPRO loss combining the PRO contrastive-plus-regularizer objective with the SFT regression (Eq.~\ref{eq:rpro_loss}):
\begin{equation}
\loss_{\text{RPRO}}(\theta) = \lambda_{\text{PRO}} \cdot \loss_{\text{PRO}}(\theta) + \lambda_{\text{SFT}} \cdot \loss_{\text{SFT}}(\theta).
\end{equation}

\section{User Study on the Data-Collection System}
\label{app:user_study}

\begin{figure}[t]
    \centering
    \includegraphics[width=0.6\linewidth]{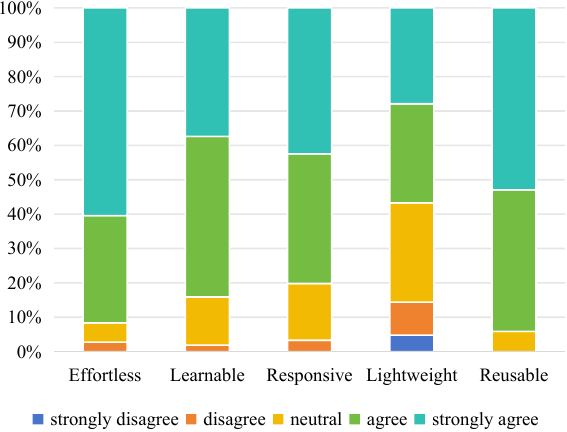}
    \caption{\textbf{User study on the teleoperated data-collection system.} Stacked-bar distribution of 5-point Likert responses across the five evaluation dimensions. Larger \emph{agree}/\emph{strongly agree} (top, green/teal) regions indicate more favorable ratings.}
    \label{fig:user_study}
\end{figure}

To assess whether our teleoperated human-in-the-loop data-collection system (Sec.~\ref{sec:data_collection}) is also \emph{usable} from the operator's perspective---and not merely effective in producing useful preference pairs for training---we conducted a small subjective user study with the data-collection operators who participated in our experiments.

\paragraph{Protocol.}
After completing several full data-collection sessions across the four tasks (\textsc{Pack}, \textsc{Cap}, \textsc{USB}, \textsc{Case}), each operator was asked to rate the system along five dimensions on a standard 5-point Likert scale (\emph{strongly disagree}, \emph{disagree}, \emph{neutral}, \emph{agree}, \emph{strongly agree}). All five items are phrased so that ``agree'' consistently corresponds to a more favorable evaluation, which avoids reverse-coding when aggregating responses. The five items, together with the one-word labels used in Fig.~\ref{fig:user_study}, are:
\begin{itemize}[leftmargin=1.2em]
    \item \textbf{Effortless} -- ``Collecting demonstrations with this system is physically effortless.''
    \item \textbf{Learnable} -- ``I was able to learn this system quickly and start collecting usable data with little training.''
    \item \textbf{Responsive} -- ``I can trigger a rollback promptly the moment I sense an upcoming failure.''
    \item \textbf{Lightweight} -- ``Resetting between trials (or after a failure) is quick and requires little extra effort.''
    \item \textbf{Reusable} -- ``I would willingly use this system again for collecting data on future tasks.''
\end{itemize}

\paragraph{Results.}
The response distributions are summarized in Fig.~\ref{fig:user_study}. Across all five dimensions the combined \emph{agree} and \emph{strongly agree} shares dominate the responses---reaching roughly $80\%$ or above for \emph{Effortless}, \emph{Learnable}, \emph{Responsive}, and \emph{Reusable}, and remaining the majority (around $56\%$) even for \emph{Lightweight}, where only about $14\%$ of responses are \emph{disagree} or \emph{strongly disagree} and the rest are neutral. Beyond producing the preference pairs that drive RPRO training, the data-collection system is therefore perceived by its users as low-effort, quick to learn, responsive to rollback, and worth reusing.

\section{Statistical Significance Analysis}
\label{app:significance}

This appendix collects the formal statistical analyses underlying the significance claims in the main text (Table~\ref{tab:main_results} and \S\ref{sec:results}). We report (i)~Wilson 95\% confidence intervals for every per-cell success rate, (ii)~Cochran--Mantel--Haenszel (CMH) tests stratified by task and base policy with Mantel--Haenszel pooled odds ratios, (iii)~a stratified sign test on the directional consistency of RPRO's advantage, and (iv)~the analogous CMH analysis for the loss-component ablations on \textsc{Pack}.

\subsection{Test setup, assumptions, and multiplicity control}
\label{app:sig_setup}

\paragraph{Data and independence.}
For each combination of base policy (PI0, PI0.5), task (\textsc{Pack}, \textsc{Cap}, \textsc{USB}, \textsc{Case}), and method, we evaluate $n{=}100$ independent rollouts with randomized initial placements. Each cell is trained with 3 random seeds; the per-cell counts are obtained by pooling the rollouts across seeds. Independence holds both within cells (fresh randomized placement per rollout) and across cells (separate fine-tuned checkpoints).

\paragraph{Test family.}
We use binomial models throughout: each cell count $X \sim \mathrm{Binomial}(n=100, p)$. Per-cell success rates are summarized by Wilson 95\% confidence intervals. Our primary cross-stratum analysis is the Cochran--Mantel--Haenszel (CMH) test with Mantel--Haenszel pooled odds ratios, and directional consistency is further assessed by a one-sided sign test over strata.

\paragraph{Multiplicity correction.}
Across the 4 pairwise CMH families (one per baseline), we interpret the CMH $p$-values under a Bonferroni correction at level $\alpha/4 = 0.0125$; all four CMH $p$-values reported below remain $< \alpha/4$ even under this stricter level.

\paragraph{Reporting conventions.}
All $p$-values below are one-sided unless otherwise stated. CMH $\chi^2$ statistics are computed without continuity correction (the standard Mantel--Haenszel score statistic). Wilson 95\% intervals use $z_{0.975}=1.96$. Mantel--Haenszel pooled odds-ratio confidence intervals use a Robins--Breslow--Greenland-type Wald approximation~\citep{robins1986general} on $\ln \widehat{OR}_{MH}$.

\subsection{Cochran--Mantel--Haenszel test, stratified by task and base policy}
\label{app:sig_cmh}

To pool evidence across the 8 task--base strata, we adopt the Cochran--Mantel--Haenszel (CMH) test, which is the standard tool for stratified $2{\times}2{\times}K$ binary outcomes and gains statistical power by combining directionally consistent evidence across strata~\citep{cochran1954some, mantel1959statistical}. We treat each task--base combination as one of $K{=}8$ strata of $2{\times}2$ tables (RPRO vs.\ baseline; success vs.\ failure; $n_R{=}n_B{=}100$ in every stratum). Let $a_k$ denote the number of RPRO successes in stratum $k$, $b_k = 100 - a_k$ failures, $c_k$ baseline successes, $d_k = 100 - c_k$ baseline failures, and $N_k = 200$. Under the null hypothesis of equal success probabilities within each stratum, the conditional mean and variance of $a_k$ given the marginals are
\begin{equation}
E_k = \tfrac{(a_k+b_k)(a_k+c_k)}{N_k} = \tfrac{a_k+c_k}{2}, \qquad
V_k = \tfrac{(a_k+b_k)(c_k+d_k)(a_k+c_k)(b_k+d_k)}{N_k^2(N_k-1)} = \tfrac{(a_k+c_k)(b_k+d_k)}{796},
\end{equation}
and the CMH score statistic is
\begin{equation}
\chi^2_{\text{CMH}} \;=\; \frac{\bigl(\sum_k(a_k - E_k)\bigr)^2}{\sum_k V_k} \;\overset{H_0}{\sim}\; \chi^2_1,
\end{equation}
which we convert to a one-sided $p$-value by halving (the observed direction matches $H_1: p_R > p_B$ in every stratum). The Mantel--Haenszel pooled odds ratio is $\widehat{OR}_{MH} = \sum_k a_k d_k / N_k \big/ \sum_k b_k c_k / N_k$, with 95\% CIs from the Robins--Breslow--Greenland-type Wald approximation on $\ln \widehat{OR}_{MH}$~\citep{robins1986general}.

Table~\ref{tab:cmh} summarizes the four pairwise CMH analyses. Across all four baselines, RPRO is significantly superior at $p < 10^{-3}$ even under a Bonferroni correction across the four families ($\alpha/4 = 0.0125$), with pooled odds ratios uniformly bounded away from 1 (lower 95\% CI $> 1.4$ in every comparison).

\begin{table}[h]
\centering
\small
\caption{\textbf{Cochran--Mantel--Haenszel tests for RPRO vs.\ each baseline, stratified by task ($\{\textsc{Pack}, \textsc{Cap}, \textsc{USB}, \textsc{Case}\}$) and base policy ($\{\textsc{PI0}, \textsc{PI0.5}\}$).} $K{=}8$ strata, $n{=}100$ per cell.}
\label{tab:cmh}
\vspace{6pt}
\setlength{\tabcolsep}{6pt}
\begin{tabular}{lcccc}
\toprule
Comparison & $\sum_k(a_k - E_k)$ & $\widehat{OR}_{MH}$ (95\% CI) & $\chi^2_{\text{CMH}}$ & $p_{\text{1-sided}}$ \\
\midrule
RPRO vs.\ DAgger          & $47.5$ & $4.58\;[3.06,\,6.87]$ & $63.13$ & $<10^{-15}$ \\
RPRO vs.\ DAgger-Buffered & $27.0$ & $2.92\;[1.92,\,4.45]$ & $26.80$ & $1.1{\times}10^{-7}$ \\
RPRO vs.\ PI0.6*          & $18.0$ & $2.24\;[1.45,\,3.47]$ & $13.88$ & $9.7{\times}10^{-5}$ \\
RPRO vs.\ TPO             & $19.0$ & $2.33\;[1.51,\,3.59]$ & $15.23$ & $4.8{\times}10^{-5}$ \\
\bottomrule
\end{tabular}
\end{table}

\subsection{Stratified sign test on directional consistency}
\label{app:sig_sign}

We further check that RPRO's advantage is \emph{directionally consistent} across strata (i.e., not driven by one or two large gaps). For each of the 8 strata, we compare RPRO to the strongest baseline within that stratum:

\begin{itemize}[leftmargin=1.2em,itemsep=1pt,topsep=1pt]
    \item Strict wins ($\text{SR}_{\text{RPRO}} > \max_b \text{SR}_b$): 8 strata---\textsc{PI0}/\textsc{Pack} ($99$ vs.\ PI0.6*~$94$), \textsc{PI0}/\textsc{Cap} ($98$ vs.\ TPO~$94$), \textsc{PI0}/\textsc{USB} ($92$ vs.\ PI0.6*~$90$), \textsc{PI0}/\textsc{Case} ($93$ vs.\ PI0.6*~$90$), \textsc{PI0.5}/\textsc{Pack} ($99$ vs.\ TPO~$95$), \textsc{PI0.5}/\textsc{Cap} ($99$ vs.\ PI0.6*~$94$), \textsc{PI0.5}/\textsc{USB} ($95$ vs.\ PI0.6*~$93$), \textsc{PI0.5}/\textsc{Case} ($93$ vs.\ TPO~$90$).
    \item Ties: 0 strata.
    \item Strict losses: 0.
\end{itemize}
A one-sided sign test of $H_1: \Pr(\text{RPRO is strictly best}) > 0.5$ yields $p = 0.5^{8} = 3.9{\times}10^{-3}$, below $0.05$, confirming that the CMH-based aggregate significance reported in \S\ref{app:sig_cmh} is not an artifact of a single dominant stratum but reflects a directionally consistent advantage.

\subsection{Loss-component ablations on \textsc{Pack} (Q3 \& Q4)}
\label{app:sig_ablation}

For the loss-component ablation reported in Fig.~\ref{fig:training_curves}(b), we evaluate $n{=}100$ rollouts under both in-distribution (ID) and out-of-distribution (OOD) initial-condition strata on \textsc{Pack}. Table~\ref{tab:ablation_cmh} reports the CMH analyses pooled across the ID/OOD strata ($K{=}2$). All four one-sided CMH $p$-values are below $6.0{\times}10^{-3}$ (raw), and remain significant at $\alpha = 0.05$ after Bonferroni correction over the 4 ablation families.

\begin{table}[h]
\centering
\small
\caption{\textbf{CMH tests for the loss-component ablation on \textsc{Pack}, stratified by ID vs.\ OOD initial conditions} ($K{=}2$ strata, $n{=}100$ per cell). $\widehat{OR}_{MH}$ confidence intervals use the Robins--Breslow--Greenland-type Wald approximation; the confidence interval for RPRO vs.\ DPO is omitted because the table contains very small baseline success counts (5/100), for which the $\ln\widehat{OR}_{MH}$ Wald interval is unreliable---the score-based $p$-value remains valid and is reported.}
\label{tab:ablation_cmh}
\vspace{6pt}
\setlength{\tabcolsep}{6pt}
\begin{tabular}{lcccc}
\toprule
Comparison & $\sum_k(a_k - E_k)$ & $\widehat{OR}_{MH}$ (95\% CI) & $\chi^2_{\text{CMH}}$ & $p_{\text{1-sided}}$ \\
\midrule
RPRO vs.\ SFT      & $10.0$ & $3.81\;[1.61,\,8.99]$  & $9.27$  & $1.2{\times}10^{-3}$ \\
RPRO vs.\ DPO+SFT  & $14.0$ & $4.82\;[2.23,\,10.40]$ & $16.05$ & $3.1{\times}10^{-5}$ \\
RPRO vs.\ PRO      & $\phantom{1}8.0$ & $2.27\;[1.20,\,4.31]$  & $6.30$  & $6.0{\times}10^{-3}$ \\
RPRO vs.\ DPO      & $83.5$ & $\sim 1.4{\times}10^{2}$ (CI omitted)\footnotemark[2] & $278.4$ & $<10^{-50}$ \\
\bottomrule
\end{tabular}
\end{table}
\footnotetext[2]{The Mantel--Haenszel pooled odds-ratio point estimate is $16823/123 \approx 136.8$; with $b_k c_k$ counts as small as $9\cdot 5=45$, the RBG Wald interval on $\ln \widehat{OR}_{MH}$ is unreliable, so we report only the score-based $\chi^2_{\text{CMH}}$ and its one-sided $p$-value, which do not require a normal approximation on $\ln \widehat{OR}$.}

\paragraph{Reading these tables.}
The pooled CMH analysis confirms that every component of RPRO contributes a statistically detectable improvement: \emph{(a)~SFT term}---adding the SFT term to PRO (i.e., RPRO vs.\ PRO) improves \textsc{Pack} success at $p \approx 6\times10^{-3}$ (CMH); \emph{(b)~Contrastive PRO term}---adding the PRO contrastive term to SFT (i.e., RPRO vs.\ SFT) improves \textsc{Pack} success at $p \approx 1.2\times10^{-3}$ (CMH), with the gap being driven primarily by the OOD stratum ($z=3.01$, $p\approx10^{-3}$ at the cell level); \emph{(c)~Proximal regularizer}---replacing PRO with plain DPO (i.e., RPRO vs.\ DPO+SFT) degrades \textsc{Pack} success at $p \approx 3.1\times10^{-5}$ (CMH), and removing both the proximal regularizer and the SFT term (i.e., RPRO vs.\ DPO) collapses success rates to $13\%$/$5\%$ at $p<10^{-50}$. Together these isolate the contribution of each loss component and substantiate the qualitative claims in \S\ref{sec:results}.

\end{document}